\documentclass[lettersize,journal]{IEEEtran}

\usepackage{amsmath,amsfonts}
\usepackage{array}
\usepackage{textcomp}
\usepackage{stfloats}
\usepackage{url}
\usepackage{verbatim}
\usepackage{graphicx}
\usepackage{amssymb}
\usepackage{hyperref}
\usepackage{graphicx}
\usepackage{amsmath}
\usepackage{longtable}
\usepackage{algorithm} 
\usepackage{algpseudocode} 
\usepackage{mathrsfs}
\usepackage{subcaption}
\usepackage{mathtools}
\usepackage{pifont}
\usepackage{color}
\usepackage{lineno}
\usepackage{makecell} 
\usepackage{pdflscape}
\usepackage{adjustbox}
\usepackage[utf8]{inputenc}
\usepackage{tabularx}
\usepackage{blindtext}
\usepackage{longtable}
\usepackage{lscape}
\usepackage{setspace}
\usepackage{notoccite} 
\usepackage{lscape} 
\usepackage{mwe}
\usepackage{booktabs}
\usepackage{amsthm}

\theoremstyle{definition}

\theoremstyle{definition}

\newcommand{\RNum}[1]{\lowercase\expandafter{\romannumeral #1\relax}}
\newcommand{\RNumU}[1]{\uppercase\expandafter{\romannumeral #1\relax}}
\usepackage[numbers]{natbib}
\usepackage{balance}
\begin{document}
\title{Enhancing Imbalance Learning: A Novel Slack-Factor Fuzzy SVM Approach}
\author{M. Tanveer{$^*$}, \IEEEmembership{Senior Member,~IEEE}, Anushka Tiwari, Mushir Akhtar, \IEEEmembership{Graduate Student Member,~IEEE}, C.T. Lin, \IEEEmembership{Fellow,~IEEE}  
\thanks{\noindent $^*$Corresponding Author\\
    M. Tanveer, Anushka Tiwari, and Mushir Akhtar are with the Department of Mathematics, Indian Institute of Technology Indore, Simrol, Indore, 453552, India (e-mail: mtanveer@iiti.ac.in, anushkatiwari9911@gmail.com, and phd2101241004@iiti.ac.in). \\
    Chin-Teng Lin is with School of Computer Science, Human-Centric AI Centre, University of Technology Sydney, Australia (e-mail: Chin-Teng.Lin@uts.edu.au).
    }}

\maketitle

\begin{abstract}
In real-world applications, class-imbalanced datasets pose significant challenges for machine learning algorithms, such as support vector machines (SVMs), particularly in effectively managing imbalance, noise, and outliers. Fuzzy support vector machines (FSVMs) address class imbalance by assigning varying fuzzy memberships to samples; however, their sensitivity to imbalanced datasets can lead to inaccurate assessments. The recently developed slack-factor-based FSVM (SFFSVM) improves traditional FSVMs by using slack factors to adjust fuzzy memberships based on misclassification likelihood, thereby rectifying misclassifications induced by the hyperplane obtained via different error cost (DEC). Building on SFFSVM, we propose an improved slack-factor-based FSVM (ISFFSVM) that introduces a novel location parameter. This novel parameter significantly advances the model by constraining the DEC hyperplane's extension, thereby mitigating the risk of misclassifying minority class samples. It ensures that majority class samples with slack factor scores approaching the location threshold are assigned lower fuzzy memberships, which enhances the model's discrimination capability. Extensive experimentation on a diverse array of real-world KEEL datasets demonstrates that the proposed ISFFSVM consistently achieves higher F1-scores, Matthews correlation coefficients (MCC), and area under the precision-recall curve (AUC-PR) compared to baseline classifiers. Consequently, the introduction of the location parameter, coupled with the slack-factor-based fuzzy membership, enables ISFFSVM to outperform traditional approaches, particularly in scenarios characterized by severe class disparity. The code for the proposed model is available at \url{https://github.com/mtanveer1/ISFFSVM}.
\end{abstract}


\section{Introduction}
\IEEEPARstart{M}{achine} learning has revolutionized numerous fields by enabling the development of algorithms that can learn from data and make accurate predictions. Its applications span a wide range of domains, such as matrix games \cite{izgi2023machine}, signal propagation \cite{ji2023signal}, analyzing criminal networks \cite{lopes2022machine}, diagnosis of Alzheimer's disease \cite{tanveer2020machine, tanveer2024ensemble, tanveer2024fuzzy} and many others.
\par
Support vector machines (SVMs) \cite{cortesc1995support, 10685140} are one such machine learning algorithm that designed to find an optimal margin that separates data points into different classes. This is accomplished by solving a convex quadratic programming problem, which identifies two hyperplanes that maximize the margin between the classes.
SVMs employ the principle of structural risk minimization (SRM) \cite{akhtar2024advancing}, which aims to prevent overfitting by minimizing the squared norm of the weight vector, effectively balancing the complexity of the model and its performance on training data. However, SVMs traditionally use the hinge loss function to measure classification errors. While hinge loss is effective in maximizing the margin, it is sensitive to outliers due to its unbounded nature \cite{akhtar2024gl, quadir2024enhancing}.
Furthermore, SVMs exhibit suboptimal performance on imbalanced datasets, where one class significantly outnumbers the other, as they tend to favor the majority class, treating minority class instances as noise. This renders SVMs inefficient for handling class imbalance, a common occurrence in real-world applications such as breast cancer diagnosis \cite{kumari2024diagnosis}, fraud detection \cite{krawczyk2016learning}, and so forth.
\par
To address imbalanced datasets, two primary methodologies exist: data-level and algorithm-level approaches \cite{krawczyk2016learning}. Data-level methods involve preprocessing techniques to balance class sizes, such as oversampling \cite{mathew2017classification}, undersampling \cite{mohammed2020machine}, and various resampling strategies \cite{more2016survey}. Oversampling introduces additional instances to the minority class, while undersampling reduces instances from the majority class. Algorithm-level methods, on the other hand, modify the training algorithm without altering the dataset. Techniques in this category include boundary-shifting methods \cite{huang2023neural}, cost-sensitive learning \cite{thai2010cost}, threshold adjustment strategies \cite{yu2015support}, and scaling kernel-based methods \cite{zhang2014imbalanced}. Cost-sensitive learning, a powerful algorithm-level approach, assigns different weights to data samples based on their class. For instance, \citet{veropoulos1999controlling} introduced the different error cost (DEC) model, where the imbalance ratio ($IR$) determines the weight of minority class samples relative to majority class samples. While the DEC model improves over standard SVM, it remains sensitive to noise and outliers.
\par
The fuzzy support vector machine (FSVM) \cite{lin2002fuzzy} enhances the standard SVM framework by incorporating a fuzzy membership (FM) function, which assigns lower weights to noisy and outlier samples. This adjustment aims to mitigate the impact of such anomalies on the overall classification performance. The FM function effectively reduces the influence of data points deemed less reliable, thereby refining the decision boundary established by the SVM. In the classical FSVM-CIL \cite{batuwita2010fsvm}, three types of tactics are applied to describe FM functions, i.e., the separation between data points and the obtained decision boundary, the separation between data points and the pre-estimated decision boundary, and the separation between data points and their own class center. The FM functions based on these three strategies have an issue of misclassification by the approximated and computed decision hyperplane. For instance, in Fig. \ref{fig:1a}, although points $A$ and $B$ are equidistant from the ideal hyperplane, point $A$ is more critical for hyperplane construction than point $B$ \cite{ren2023slack}. To avoid misclassifying $A$, the hyperplane is pushed left, ensuring proper separation of the majority class, which makes $A$ more important. Furthermore, the class imbalance is another factor for the incorrect display of the importance of the samples by FM values.
\par
To overcome these limitations, slack-factor-based fuzzy support vector machine (SFFSVM) \cite{ren2023slack} was developed. The SFFSVM model introduces the concept of a slack factor, which is used to define a FM function. The primary objective of incorporating slack factor values in the SFFSVM model is to move the decision hyperplane $M_{dec}$, initially derived from the DEC model, towards the optimal hyperplane $M^*$ (as depicted in Fig. \ref{fig:1b}).
Higher slack factor values indicate a higher probability that the sample is either an outlier or contains noise. Consequently, such samples are assigned lower FM values to reduce their impact on the decision boundary. This approach allows the model to maintain a more robust decision surface by diminishing the influence of less reliable data points. The SFFSVM model also has a drawback while defining FM values for the majority class samples. The FM function, defined for the majority class samples, assigns high FM values to samples misclassified by the DEC model hyperplane, provided their slack factor value is less than 2 (see Eq. (\ref{eq:12})). However, \(2\) is not always the optimal choice, as shifting \(M_{dec}\) to the right must be done carefully to prevent the misclassification of correctly classified minority class samples.
\par
Taking motivation from the prior research, in this paper, we propose an improved slack-factor-based fuzzy support vector machine (ISFFSVM) model, which introduces a new parameter, named as the location parameter, denoted by $a$. The advantage of introducing the location parameter $a$ is that the number of minority class samples that will be misclassified after shifting the $M_{dec}$ hyperplane depends on the parameter $a$. Consequently, the proposed ISFFSVM model is more efficient for class imbalance learning as it attempts to correctly classify a larger number of minority samples. Moreover, Fig. \ref{fig:3.2} demonstrates that the decision boundary of the proposed ISFFSVM model is superior to the SFFSVM model in classifying minority class samples.\\
The key highlights of the paper can be encapsulated as follows:
\begin{enumerate}
    \item We introduce the ISFFSVM, which combines slack factor-based fuzzy membership with a novel location parameter, providing a robust solution for managing severe class disparity, reducing misclassification rates, and enhancing overall model performance.
    \item The introduction of the location parameter provides a significant advancement by constraining the DEC hyperplane's extension, mitigating the risk of misclassifying minority class samples, and optimizing the overall decision boundary adjustment process for better handling of class imbalances.
    \item We demonstrate that the proposed ISFFSVM outperforms baseline classifiers on diverse real-world KEEL datasets, achieving higher F1-scores, Matthews correlation coefficients (MCC), and area under the precision-recall curve (AUC-PR), effectively addressing class imbalance challenges.
\end{enumerate}

The remaining sections of the paper are organized as follows: The approaches for class imbalance learning available in the literature are discussed in Section 2. The proposed method is explained in Section 3. In Section 4, experimental comparisons are thoroughly presented and Section 5 includes conclusions with future work.

\section{Related Work}
This section briefly explains the related work on the imbalanced datasets. Over the past few decades, numerous algorithm-level techniques have been proposed. Some of them are given below:
\subsection{Different Error Cost (DEC)}
In DEC \cite{veropoulos1999controlling} model, a variant of SVM, an imbalance ratio ($IR$) has been introduced to deal with class imbalance problems. The minority class samples' misclassification cost is $IR$ in the DEC. It is better than SVM on the imbalanced dataset due to the involvement of the $IR$. The optimization problem of DEC is as follows:

\begin{align}
\label{eq:1}
\underset{w,b,\xi_{i}}{\min }\,& \frac{1}{2}w^2+\zeta^+\sum_{x\in X^+}^{|X^+|}\xi _{i} +\zeta^-\sum_{x\in X^-}^{|X^-|}\xi _{i} \\
\text{s.t.} \hspace{0.2cm}  &  y_i (w^T x_{i} + b) \geq 1-\xi_{i}, i= 1,2,\hdots,N, \nonumber \vspace{0.3cm}\\ 
& \xi_{i} \geq 0, i=1,2,\hdots,N, \nonumber
\end{align}
where $\zeta^-=\zeta, \zeta^+=\zeta*IR,$ and $\zeta~\text{is regularization parameter}$.\\
DEC model is increasing the membership of the minority class by multiplying the corresponding slack factors with $IR$. In this way, it assigns different FM values to the majority and minority class samples. However, like SVM, DEC model is sensitive to noise and outliers as it cannot distinguish them properly. 

\subsection{Slack Factor Based Fuzzy Support Vector Machine (SFFSVM)}
We discuss the formulation of SFFSVM \cite{ren2023slack} in this section.
SFFSVM defines slack-factor-based FM which helps to define different FM values for the misclassified and correctly classified samples based on their importance. In this way, it reduces the impact of imbalance and noise or outliers on the FM values.
\begin{align}\label{eq:2} \underset{w,b,\xi_{i}}{\min }\,& 
 \hspace{0.1cm}\frac{1}{2}w^2+\zeta\sum_{i= 1}^{N}\xi _{i}  \\ \text{s.t.} \hspace{0.2cm}  &  y_i (w^T x_{i} + b) \geq 1-\xi_{i}, i= 1,2,\hdots,N,
 \nonumber \\ & \xi_{i} \geq 0, i=1,2,...,N.\ \nonumber
\end{align}
The decision boundary $M_{svm} = \{x|\tilde{{w}}^Tx+\tilde{{b}} \hspace{0.6mm} \cap \hspace{0.6mm} x \in \mathbb{R}^d \}$ can be constructed by optimizing Eq. (\ref{eq:2}) on the dataset $X=\{(x_{i},y_{i})\}_{i=1}^{N}$, where $\tilde{w}$ and $\tilde{b}$ represent the best possible outcome. After obtaining $M_{svm}$ on $X$, we determine the slack factor value of a sample $x_{i}$ using the hinge loss function\\
$$\xi_{i}=\max (0,1-y_{i}({w}^Tx_{i}+{b})).$$
A slack-factor-based FM is defined below based on the observation that the likelihood of misclassifying the related sample increases with the size of the slack factor values.
\begin{equation}
\label{eq:10} 
    \psi_ {x_{i}}=\left\lbrace \begin{array}{lr} 1, &\,x_{i} \in T,\\  e^{\left(-\mu\xi_{i}\right)}, &x_{i}\in F.\end{array}\right.
\end{equation}
Here,
\begin{align*}
 T =&\left\lbrace x_{i}|x_{i} \in X \cap (0 \leq\xi_{i} < 1)\right\rbrace ,\\
F =&\left\lbrace x_{i}|x_{i} \in X \cap (\xi_{i} \geq 1) \right\rbrace,
\end{align*}
 and $\mu$ is the smoothness parameter which determines how smoothly the DEC hyperplane moves.

\begin{figure*}
\centering
\subcaptionbox{\label{fig:1a}} {%
\includegraphics[width=0.52\textwidth,keepaspectratio]{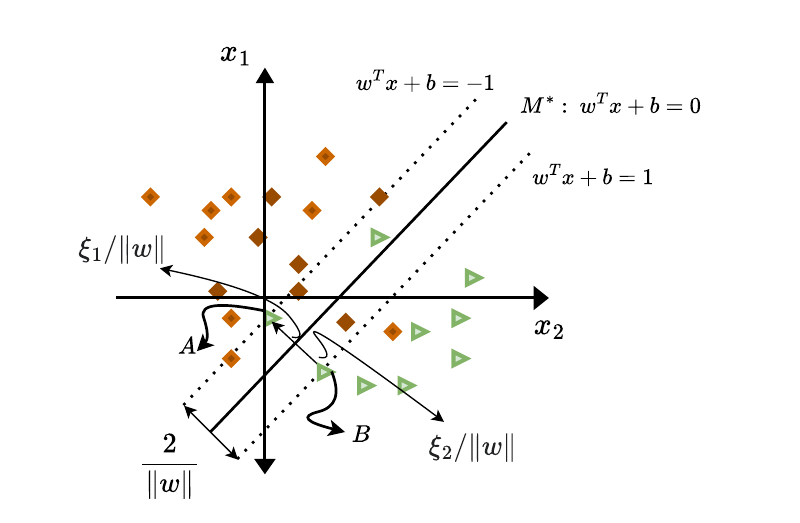}}
\hfill
\subcaptionbox{\label{fig:1b}} { %
\includegraphics[width=0.43\textwidth,keepaspectratio]{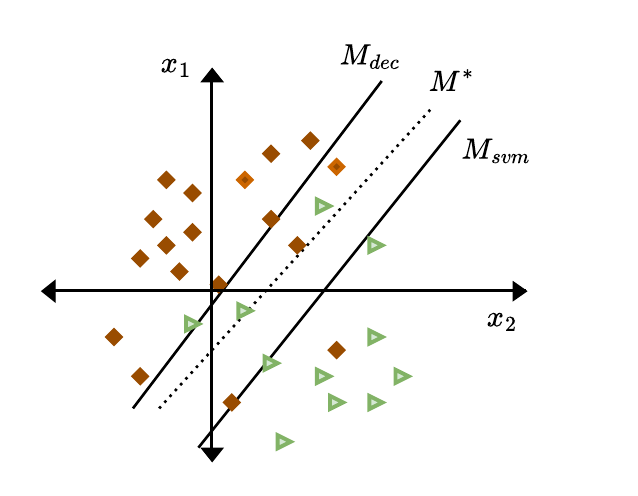}}
      \caption{Decision boundary visualization for SFFSVM and proposed ISFFSVM. (a) Illustration of decision hyperplane and slack factors for SFFSVM model. Points $A$ and $B$ are equidistant from the ideal hyperplane \(M^*\), while differing in their importance for hyperplane construction. (b) Improved decision boundary of proposed ISFFSVM model. Introduction of location parameter $a$ ensures correctly classified minority samples are less likely to be misclassified.}
    \label{fig:Decision_boundary}
 \end{figure*}

\begin{figure}
\centering
\includegraphics[width=0.5 \textwidth,height=7cm]{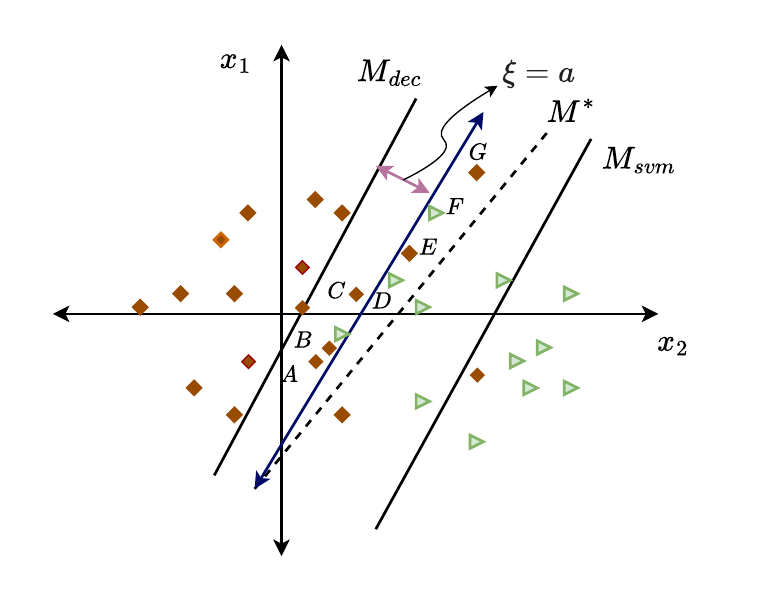}
\caption{Illustrates the impact of the location parameter $a$ on membership values of majority class samples. Points $A$, $B$, and $C$ with slack factor values less than $a$ are assigned one membership value, whereas points $E$ and $G$ with slack factor values between $a$ and $2$ are assigned lower membership values.}
\label{fig:3.1}
\end{figure}

SFFSVM defines a new FM function based on the above slack-factor-based FM function. It uses DEC to find the slack factor values and then define FM values for majority and minority data samples.
Conveniently, suppose that $M^*$ and $M_{dec}$ denote the most ideal decision hyperplane on $X=\lbrace X^+ \cup X^-\rbrace$ and a decision hyperplane derived by instructing DEC on $X$, respectively. Depending on the values of $\xi^+(\text{or}\hspace{0.1cm} \xi^-)$ of $M_{dec}$ can divide $X^+(X^-)$ into $T^+(T^-)$ and $F^+(F^-)$. Note that, we apply superscripts $``-"$ and $``+"$ to represent the majority and minority class samples, respectively. For instance, $T^+$ denotes the collection of minority class samples correctly classified with a slack factor greater than or equal to zero and less than one.

The following FM values are set for the minority class samples:
\begin{equation}
\label{eq:11} 
\psi^+_{x_{i}}=\left \lbrace \begin{array}{lr} 2/(e^{\mu\xi_{i}}+1), &x_{i}\in T^+,\\0,&x_{i}\in F^+.\end{array}\right.
\end{equation}
Eq. (\ref{eq:11}) depicts that the FM value for correctly classified minority class samples decreases exponentially as the slack factor value increases. At the same time, FM value for wrongly classified minority class samples is zero.

The following FM values are set for the majority class samples:
\begin{align}
\label{eq:12} 
\psi^-_{x_{i}}=\left\lbrace \begin{array}{lr} e^{-\mu\xi_{i}},&x_{i} \in \left\lbrace x \mid x\in F^-\cap\xi(x)\geq2 \right\rbrace, \\
1,&x_{i}\in \left\lbrace x \mid x\in F^-\cup T^- \cap \xi(x) < 2 \right\rbrace.\end{array} \right.
\end{align}
Here, the FM value for majority class samples is set to one when their slack factor is less than $2$, and decreases exponentially when the slack factor is greater than or equal to $2$.

Implementation of the above memberships in the DEC model can be expressed as follows:
\begin{align}\label{eq:13}  \underset{w,b,\xi_{i}}{\min }\,& \hspace{0.1cm}\frac{1}{2}w^2+\zeta^+\sum_{x\in X^+}^{|X^+|}\psi^+_{x_{i}}\xi _{i} +\zeta^-\sum_{x\in X^-}^{|X^-|}\psi^-_{x_{i}}\xi _{i}  \\ \text{s.t.} \hspace{0.2cm}  &  y_i (w^T x_{i} + b) \geq 1-\xi_{i}, i= 1,2,\hdots,N, \vspace{0.3cm}
\nonumber \\ & \xi_{i} \geq 0, i=1,2,\hdots,N, \nonumber
\end{align}


where $\zeta^-=\zeta \text{ and }\zeta^+=\zeta*IR.$\\

SFFSVM considers the concept of slack factor, which helps to resolve  the issue of assigning equal membership to points, $A$ and $B$ as depicted in Fig. \ref{fig:1a}. Moreover, slack-factor-based FM functions are less affected by class imbalance problems since they employ DEC to find the FM values.
\section{Proposed Work}
FSVM-CIL \cite{batuwita2010fsvm} has an issue of misclassification by the approximated and computed decision hyperplane. For instance, in Fig. \ref{fig:1a}, points $A$ and $B$ are equally far from the ideal hyperplane $M^*$, however, $A$ is more crucial in constructing the hyperplane than $B$. Furthermore, the class imbalance is another factor for the incorrect display of the importance of the samples by FM values. To overcome this issue, slack-factor-based FM function has been proposed in the literature \cite{ren2023slack}. In SFFSVM, all the majority class samples wrongly classified by the DEC hyperplane, i.e., they are on the right side of the DEC hyperplane and with slack factor value less than $2$, are allocated one membership value. In other words, majority class samples between the decision boundary obtained by DEC and the supporting hyperplane $w^T_{dec}x+b_{dec}=1$ are given equal membership as the majority class samples which are correctly classified by the DEC hyperplane.

\begin{figure}
\begin{minipage}{.1\linewidth}
\centering
\subfloat[Decision surface of SFFSVM]{\label{main:a}\includegraphics[scale=0.55]{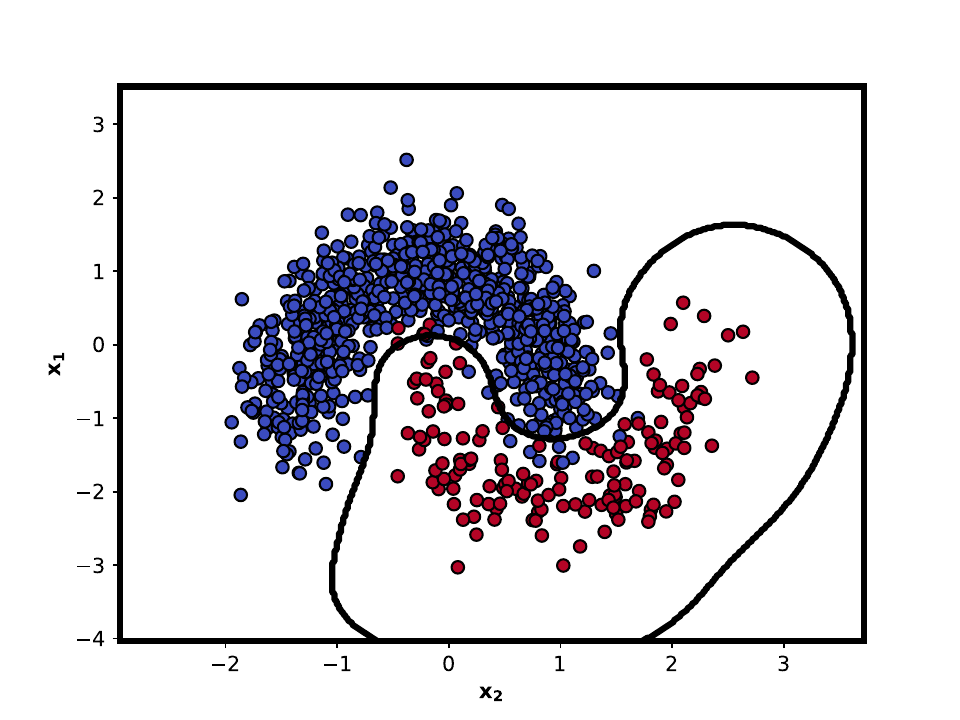}}
\end{minipage}
\par\medskip
\begin{minipage}{.3\linewidth}
\centering
\subfloat[Decision surface of ISFFSVM]{\label{main:c}\includegraphics[scale=0.55]{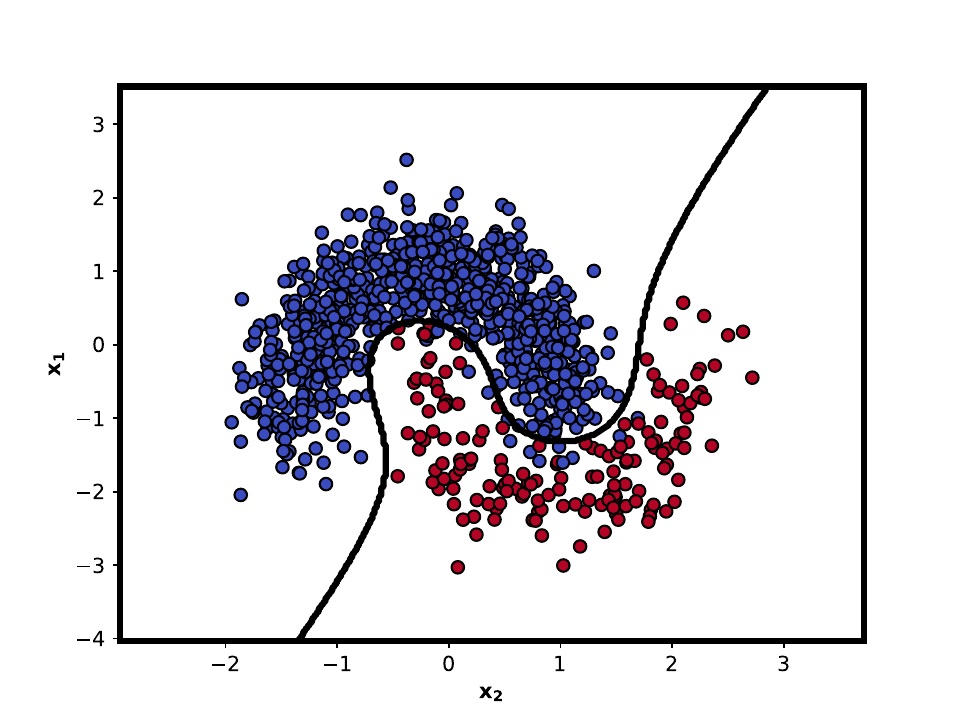}}
\end{minipage}
\par\medskip
\caption{Comparison of decision hyperplanes for SFFSVM and proposed ISFFSVM. (a) Decision surface of SFFSVM on moon dataset containing $1000$ majority class data points (blue dots) and $200$ minority class data points (red dots) (i.e., $IR = 5$). (b) Decision surface of ISFFSVM on the same dataset. The proposed ISFFSVM model demonstrates superior classification of minority class samples due to the introduction of the location parameter $a$, which adjusts membership values more effectively.}
\label{fig:3.2}
\end{figure}

In the proposed ISFFSVM model, we introduce a new parameter \(a\), given in Eq. (\ref{eq:9}), referred to as the location parameter. Unlike the SFFSVM model, we assign a membership value of one only to those majority class samples with slack factor values greater than zero and less than \(a\). For example, in Fig. \ref{fig:3.1}, consider points \(A\), \(B\), and \(C\), which have slack factor values less than \(a\) (with \(a < 2\)). In contrast, points \(E\) and \(G\) have slack factor values between \(a\) and $2$, inclusive. According to Eq. (\ref{eq:9}), points \(A\), \(B\), and \(C\) will be assigned a membership value of one, whereas points \(E\) and \(G\) will be assigned a membership value less than one. This approach ensures that correctly classified minority points \( D \) and \( F \) are less likely to be misclassified when shifting the DEC hyperplane \( M_{dec} \) to the right, as high membership values are not assigned to samples \( E \) and \( G \). Thus, more minority samples will be correctly classified, which is our priority in class imbalance learning. However, as per the SFFSVM model, $E$ and $G$ will also be assigned one membership value. Consequently, there is a very high chance that minority samples $D$ and $F$ will be misclassified which is contrary to our objective. Also, we can verify through Fig. \ref{fig:3.2} that, in comparison to SFFSVM model, the proposed ISFFSVM model successfully classifies a larger number of minority samples.

The following DEC model is employed to obtain the FM values:
\begin{align}
\label{eq:7}
\underset{w,b,\xi_{i}}{\min }\,& \frac{1}{2}w^2+\zeta^+\sum_{x\in X^+}^{|X^+|}\xi _{i} +\zeta^-\sum_{x\in X^-}^{|X^-|}\xi _{i} \\
\text{s.t.} \hspace{0.2cm}  &  y_i (w^T x_{i} + b) \geq 1-\xi_{i}, i= 1,2,\hdots,N,\vspace{0.3cm} \nonumber \\
& \xi_{i} \geq 0, i=1,2,\hdots,N. \nonumber
\end{align}
FM values set for the minority class samples are as follows:
\begin{align}
\label{eq:8} 
    \psi^+_{x_{i}}=\left\lbrace \begin{array}{lr} 2/(e^{\mu\xi_{i}}+1), &\,x_{i} \in T^+,\\ 0, &x_{i}\in F^+.\end{array}\right.
\end{align}
FM values set for the majority class samples are as follows:
\begin{align}
\label{eq:9} 
&\psi^-_{x_{i}}=
\left\lbrace \begin{array}{lr} e^{-\mu\xi_{i}},&x_{i} \in \left\lbrace x \mid x\in F^-\cap\xi(x)\geq a \right\rbrace, \\
1,&x_{i}\in \left\lbrace x \mid x\in F^-\cup T^- \cap \xi(x) < a \right\rbrace.\end{array} \right.
\end{align}

 Implementation of the above membership function in the DEC model can be expressed as follows:
\begin{align}\label{eq:10}
\underset{w,b,\xi_{i}}{\min }\,& \frac{1}{2}w^2+\zeta^+\sum_{x\in X^+}^{|X^+|}\psi^+_{x_{i}}\xi _{i} +\zeta^-\sum_{x\in X^-}^{|X^-|}\psi^-_{x_{i}}\xi _{i}  \\ \text{s.t.} \hspace{0.2cm}  &  y_i (w^T x_{i} + b) \geq 1-\xi_{i}, i= 1,2,\hdots,N,\vspace{0.3cm}
\nonumber \\ & \xi_{i} \geq 0, i=1,2,\hdots,N, \nonumber
\end{align}\\
where $\zeta^-=\zeta \text{ and }\zeta^+=\zeta*IR.$

We are tuning the parameter $a$ in the range $[1.1:0.1:2]$. It has been noted that $2$ is not always the most suitable choice for the parameter $a$; values other than $2$ are appearing as the best value for $a$. For instance, in Fig. \ref{fig:3.2} (b), the decision boundary is drawn with $a=1.3$. It can be seen that the decision boundary of the proposed ISFFSVM model is better as compared to SFFSVM model. The proposed ISFFSVM model misclassifies a lesser number of minority samples than SFFSVM model; consequently, it deals with class imbalance more competently than SFFSVM model.

Unlike DEC, the proposed ISFFSVM model assigns varying FM values to samples based on their significance. Additionally, unlike the SFFSVM model, it does not always use 2 as the threshold for assigning an FM value of one to majority class samples. With the introduction of the location parameter \(a\), the ISFFSVM model delivers more accurate FM values. For \(a = 2\), the proposed ISFFSVM reduces to the baseline SFFSVM model, maintaining consistency while offering greater flexibility for improved performance when \(a\) is adjusted.


\begin{algorithm}
  \caption{Improved Slack-Factor-Based Fuzzy Support Vector Machine}
  \begin{algorithmic}[1]
     \State Let $X\in \mathbb{R}^{N \times d}$ be the given input dataset and $Y$ contains the target labels.
     \State Obtain optimal $w$ and $b$ using Eq. (\ref{eq:7}).
      \For{$i=1,2,\cdots,N $}
      \State Calculate slack factor value for each $x_{i}$ using the formula $\xi_{i}=\max (0,1-y_{i}({w}^Tx_{i}+{b}))$.
\If{$y_{i}=1$}
 \State if $\xi_{i}<1$ put $x_{i}$ into $T^+$, otherwise into $F^+$.\\
           ~~~~~~~~Calculate FM value using Eq. (\ref{eq:8}).
\Else 
\State if $\xi_{i}<1$ put $x_{i}$ into $T^-$, otherwise into $F^-$.\\
 ~~~~~~~~~Calculate FM value using Eq. (\ref{eq:9}).
 \EndIf
 \EndFor
 \State Apply obtained FM values in Eq. (\ref{eq:10}) and calculate the optimal $w$ and $b$.
 \end{algorithmic}
\end{algorithm}

\subsection{Theoretical Justification of the Proposed ISFFSVM}
This subsection delves into the mathematical foundations and theoretical advantages of the proposed ISFFSVM model, specifically focusing on the strategic incorporation of the slack factor and location parameter. The following are the key theoretical aspects that underpin the robustness and effectiveness of the ISFFSVM model:\\
\textbf{Membership assignment:} The proposed ISFFSVM model assigns a membership value of one only to majority class samples with slack factor values between zero and \(a\) (i.e., \(0 < \xi < a\)). For example, in Fig. \ref{fig:3.1}, points \(A\), \(B\), and \(C\) with slack factor values below \(a\) are assigned a membership value of one, while points \(E\) and \(G\), with slack factor values between \(a\) and 2 (i.e., \(a \leq \xi < 2\)), receive a membership value less than one. This selective assignment allows for more precise handling of majority class samples.\\
\textbf{Impact on hyperplane adjustment:} By ensuring that only majority class samples with slack factor values less than \(a\) receive high membership values, the proposed ISFFSVM model better positions the DEC hyperplane \(M_{\text{dec}}\) when shifted to the right. This reduces the risk of misclassifying correctly classified minority class samples, such as points $D$ and $F$ (see Fig. \ref{fig:3.1}), by not assigning high membership values to points with higher slack factor values like $E$ and $G$.\\
\textbf{Effect on minority class classification:} The proposed ISFFSVM model prioritizes the correct classification of minority class samples by differentiating membership values based on the slack factor. Unlike the baseline SFFSVM model, where points $E$ and $G$ could receive a membership value of one, potentially leading to misclassification of minority class samples like $D$ and $F$, the ISFFSVM model maintains a more accurate separation between majority and minority class samples.\\
\textbf{Verification and comparison:} Fig. \ref{fig:3.2} demonstrates the decision boundaries of both the baseline SFFSVM and the proposed ISFFSVM models. The proposed ISFFSVM model shows superior classification accuracy for minority class samples due to the nuanced assignment of membership values, which optimally positions the DEC hyperplane to favor the correct classification of minority samples.

\subsection{Error Analysis for the Proposed ISFFSVM}
The proposed ISFFSVM introduces a novel approach to identifying and handling misclassifications, particularly false positives (FP) and false negatives (FN), which are critical in the context of imbalanced datasets. The model's ability to differentiate between these types of errors is rooted in the innovative use of a slack-factor-based fuzzy membership function combined with a newly introduced location parameter, \(a\).
\begin{itemize}
\item  \textbf{Identification of False Positives (FP):}
   In the ISFFSVM framework, false positives occur when majority class samples are incorrectly classified as minority class samples. The model leverages the slack factor to compute fuzzy membership values, which determine the importance of each data point in influencing the decision boundary. For majority class samples, the ISFFSVM assigns fuzzy memberships based on their slack factor values and the location parameter \(a\). Majority class samples with slack factor values below \(a\) receive a high fuzzy membership value (close to 1), while those with slack factor values greater than or equal to \(a\) are assigned lower fuzzy membership values. This mechanism effectively reduces the weight of majority class samples that are more likely to be misclassified (i.e., close to the decision boundary), thereby lowering the FP rate by minimizing their influence on the decision boundary.
\item \textbf{Identification of False Negatives (FN):}
   False negatives occur when minority class samples are incorrectly classified as majority class samples. The ISFFSVM model addresses this issue by adjusting the decision boundary to reduce the likelihood of minority class samples being misclassified. The location parameter \(a\) plays a crucial role in this process. By shifting the decision hyperplane (M\(_{\text{dec}}\)) based on the location parameter, the model ensures that correctly classified minority class samples are less likely to be misclassified as the decision boundary moves. Specifically, minority class samples with low slack factor values receive high fuzzy memberships, emphasizing their importance in defining the optimal decision boundary. This selective assignment of fuzzy memberships reduces the influence of majority class samples near the boundary and helps maintain a clearer separation, thereby decreasing the FN rate.
\item \textbf{Impact of the Location Parameter (\(a\)):}
   The introduction of the location parameter \(a\) allows the ISFFSVM to finely tune the decision boundary's position by adjusting the fuzzy membership values for majority class samples. Samples with slack factor values close to \(a\) are carefully weighted to ensure that the decision boundary remains optimal for distinguishing between minority and majority classes. By doing so, the model reduces the chance of misclassifying critical minority class samples (reducing FN) and avoids undue influence from misclassified majority class samples (reducing FP).
\item \textbf{Overall Error Reduction:}
   Through this novel integration of the slack-factor-based fuzzy membership and the location parameter, the ISFFSVM effectively minimizes both FP and FN rates. The model dynamically adjusts the membership values to maintain a robust decision boundary that is less sensitive to noise and outliers, providing superior classification performance across various imbalanced datasets. The proposed method's ability to differentiate and minimize both types of errors demonstrates its robustness and adaptability to real-world scenarios characterized by severe class imbalance.
\end{itemize}
This error analysis highlights how the proposed ISFFSVM effectively identifies and mitigates different types of misclassifications, thereby enhancing its overall classification performance and ensuring its suitability for imbalanced data applications.

\subsection{Computational Complexity of the proposed ISFFSVM}
The computational complexity and scalability of the proposed ISFFSVM model is similar to that of the baseline SFFSVM model, with an additional time cost associated with tuning the location parameter \(a\). Let \(N\) denote the number of samples in the dataset and \(d\) the number of features. The computational complexity of training the ISFFSVM model, like the SFFSVM, primarily depends on the training of the SVM with the DEC formulation. Using the Bunch-Kaufman algorithm, the complexity of training DEC is bounded between \(O(N_s^3 + dN_s^2 + dN^2)\) and \(O(dN^2)\), where \(N_s\) is the number of support vectors, and typically \(N_s \ll N\). This gives an overall complexity of \(O(dN_s^2 + dN^2 + N_s^3) = O(dN^2)\) \cite{ren2023slack}. The complexity of assigning fuzzy memberships to samples in ISFFSVM remains the same as in SFFSVM, which is \(O(N)\). This arises from the need to evaluate each sample's slack factor in order to compute its fuzzy membership value using the parameter \(a\). Therefore, the overall computational complexity of the proposed ISFFSVM model is \(O(dN^2)\), the same as the SFFSVM model. However, ISFFSVM requires an additional time cost for tuning the location parameter \(a\). This analysis highlights that the proposed ISFFSVM model maintains the same computational efficiency and scalability as the baseline SFFSVM model while offering improved performance by optimizing the decision boundary through the additional parameter \(a\).

\section{Numerical Experiments}
To evaluate the efficacy of the proposed ISFFSVM, we compared it with the following baseline models: $(1)$ Ensemble-based methods: Hashing-based Under-sampling Ensemble (HUE) \cite{ng2020hashing}; $(2)$ Fuzzy-based methods: FSVM based on centered kernel alignment (CKA-FSVM) \cite{wang2020centered} and Fuzzy SVM for class imbalance learning \cite{batuwita2010fsvm} (FSVM-CIL: FSVM-CIL-exp and FSVM-CIL-lin); $(3)$ Other category: Different Error Cost (DEC) \cite{veropoulos1999controlling}, Complement Naive Bayes (CNB) \cite{rennie2003tackling}, and Synthetic Minority Oversampling Technique SVM (SMOTE-SVM) \cite{guo2024adaptive}; $(4)$ Oversampling-based category: Polynomial fit SMOTE (PF-SMOTE) \cite{gazzah2008new}, SMOTE-Tomeklinks \cite{batista2004study} and, MWMOTE \cite{barua2012mwmote}.
For the assessment of the  models, we choose real-world datasets given by Knowledge Extraction Based on Evolutionary Learning (KEEL) \cite{derrac2015keel}. All datasets are divided into two groups: one with $IR<10$ and another with $IR \geq 10$. The experiments are performed on a machine with Python 3.7
on the system with 2 Intel Xeon processors, 128 GB
of RAM, and 4 TB of secondary storage. The hyperparameter setting for all the models is taken the same as given in \cite{ren2023slack}.
For the proposed ISFFSVM, the location parameter $a$ is chosen from the range $\left[1.1:0.1:2\right]$.
Each dataset is divided into an $80:20$ ratio for training and testing the models. We used the grid search method to tune
the hyperparameters via a five-fold cross-validation method. To reduce the variability of the results, we perform five-fold cross-validation ten times independently on each dataset.

\subsection{Evaluation Metrics}
We select the area under the precision-recall curve (AUC-PR), the Matthews correlation coefficient (MCC), and the F1-score as the evaluation metrics, as accuracy is not suitable for assessing performance under class imbalance conditions \cite{he2005over, wang2020learning, wang2022generalization}. To define the F1-score and MCC, we first provide the formulas for precision and recall. These metrics are derived from the confusion matrix, which indicates the number of samples that are correctly or incorrectly classified by the model for each class.
The formulas for precision and recall are as follows:

\[
\text{Precision}(P) = \frac{T_P}{T_P + F_P} \quad \text{and} \quad \text{Recall}(R) = \frac{T_P}{T_P + F_N},
\]
where \( TP \), \( FP \), and \( FN \) denote true positives, false positives, and false negatives, respectively.

Using these definitions, the F1-score and MCC can be computed as:
\begin{equation}
    \text{F1-score}=\frac{2 \times R \times P}{R + P},
\end{equation}

\begin{align}
\label{eq:21} 
&\text{MCC}=
&\frac{T_{P} \times T_{N}-F_{P} \times F_{N}}{\sqrt{(T_{P}+F_{P})(T_{P}+F_{N})(T_{N}+F_{P})(T_{N}+F_{N})}}.
\end{align}

The AUC-PR measures the area under the curve formed by plotting precision against recall across different thresholds.

\subsection{Experimental Analysis}
In this subsection, we provide a detailed evaluation of the proposed ISFFSVM model against several baseline models using three metrics: F1-score, MCC, and AUC-PR. We analyze the results separately for low $IR$ and high $IR$ datasets.
\subsubsection{Performance on Low $IR$ Datasets}
{The evaluation results on low $IR$ datasets, as summarized in Tables \ref{Table1} indicate that the proposed ISFFSVM model significantly outperforms the baseline models in terms of average performance across all three metrics. The detailed results for each of the low $IR$ datasets, based on F1-score, MCC, and AUC-PR, are presented in Tables S.V, S.VI, and S.VII, respectively, in the supplementary file. The proposed ISFFSVM achieves an average F1-score of $80.14\%$, surpassing the average F1-score of the two best-performing baseline models, including SFFSVM ($79.13\%$) and SMOTE-SVM ($77.68\%$) (see Table \ref{Table1}). This suggests that the ISFFSVM model maintains a better balance between precision and recall when handling moderate class imbalance. In terms of MCC, the proposed ISFFSVM achieves an average score of $69.68\%$, surpassing the average MCC of the two best-performing baseline models, SMOTE-SVM (69.01\%) and SFFSVM (68.52\%) (see Table \ref{Table1}). This indicates that the proposed ISFFSVM is more effective at capturing the correlation between true and predicted labels, reducing both false positives and false negatives. Moreover, the proposed model attains the highest average AUC-PR score of $83.38\%$, which is greater than that of baseline models like SFFSVM ($82.27\%$) and SMOTE-SVM ($81.80\%$) (see Table \ref{Table1}). These results highlight that the introduction of the location parameter and the slack-factor-based fuzzy memberships in the proposed ISFFSVM model leads to better discrimination capability and overall performance on low $IR$ datasets.

\subsubsection{Performance on High $IR$ Datasets}
{The average performance evaluation on high $IR$ datasets, as presented in Table \ref{Table4}, further confirms the superiority of the proposed ISFFSVM model over the baseline models. The detailed results for each of the high $IR$ datasets, evaluated using F1-score, MCC, and AUC-PR, are provided in Tables S.VIII, S.IX, and S.X, respectively, in the supplementary file. The average F1-score of ISFFSVM is $53.74\%$, which is higher than the average F1-scores of baseline models like SFFSVM ($52.01\%$) and  SMOTE-SVM ($50.99\%$) (see Table \ref{Table4}). This demonstrates that ISFFSVM effectively handles severe class imbalance, maintaining better performance in comparison to baseline models. Similarly, the proposed model achieves the highest average MCC score of $53.98\%$, which is greater than that of SFFSVM ($52.36\%$) and SMOTE-SVM ($52.18\%$) (see Table \ref{Table4}). The superior MCC score indicates that ISFFSVM establishes a robust decision boundary that reduces both false positives and false negatives. Additionally, the proposed ISFFSVM model achieves the highest average AUC-PR score of $58.89\%$, outperforming the best two baseline models like SFFSVM ($58\%$) and SMOTE-SVM ($57.10\%$) (see Table \ref{Table4}). The high AUC-PR score highlights ISFFSVM’s strong capability to maintain a good balance between precision and recall, even in the presence of a highly skewed class distribution.}

Overall, the proposed ISFFSVM model leverages its unique slack-factor-based fuzzy memberships and location parameter to achieve superior performance in handling class imbalance. It outperforms all the baseline models, including SFFSVM and SMOTE-SVM, in terms of all three metrics on both low and high $IR$ datasets, making it a robust and efficient solution for imbalanced classification problems.

To further validate the effectiveness of the proposed ISFFSVM model, we performed a comprehensive statistical analysis using ranking tests, the Friedman test, and the Nemenyi post hoc test. The results demonstrate that the proposed ISFFSVM model performs statistically better than the baseline models. A detailed discussion of the statistical analysis is provided in Section S.I.A of the supplementary file.}

\begin{table*}
\centering
\caption{Presents the average F1-scores, MCC scores, and AUC-PR for the proposed ISFFSVM model and baseline models on datasets with low $IR$.}
\label{Table1}
\resizebox{1\textwidth}{!}{
\tiny
\begin{tabular}{lcccccccccccc}
\hline
&DEC \cite{veropoulos1999controlling}&FSVMCIL$_{exp}$ \cite{batuwita2010fsvm}&FSVMCIL$_{lin}$
\cite{batuwita2010fsvm}&CKAFSVM \cite{wang2020centered}&MWMOTE \cite{barua2012mwmote}&PFSMOTE \cite{gazzah2008new}&SMOTETL \cite{batista2004study}&HUE \cite{ng2020hashing}&CNB \cite{rennie2003tackling}&SFFSVM \cite{ren2023slack}&SMOTE-SVM \cite{guo2024adaptive}&ISFFSVM\\
\hline
\textbf{Average F1-score} &73.85&73.6&73.34&73.02&72.53&73.43&71.8&72.83&58.42&79.13&77.68&\textbf{80.14}\\ \hline
\textbf{Average MCC} & 66.8&66.08&65.71&66.27&64.44&66.15&63.62&64.89&45.97&68.52&69.01&\textbf{69.68}\\\hline
\textbf{Average AUC-PR} & 78.34&78.29&78.32&78.75&75.59&77.18&75.01&79.43&57.25&82.27&81.80&\textbf{83.38}\\	\hline
\end{tabular}}
\end{table*}
 
\begin{table*}
\centering
\caption{Presents the average F1-scores, MCC scores, and AUC-PR for the proposed ISFFSVM model and baseline models on datasets with high $IR$.}
\label{Table4}
\resizebox{1\textwidth}{!}{
\tiny
\begin{tabular}{lcccccccccccc}
\hline
&DEC \cite{veropoulos1999controlling}&FSVMCIL$_{exp}$ \cite{batuwita2010fsvm}&FSVMCIL$_{lin}$
\cite{batuwita2010fsvm}&CKAFSVM \cite{wang2020centered}&MWMOTE \cite{barua2012mwmote}&PFSMOTE \cite{gazzah2008new}&SMOTETL \cite{batista2004study}&HUE \cite{ng2020hashing}&CNB \cite{rennie2003tackling}&SFFSVM \cite{ren2023slack}&SMOTE-SVM \cite{guo2024adaptive}&ISFFSVM\\
\hline
\textbf{Average F1-score} &47.24&43&40.47&41.74&43.15&46.82&45.03&38.8&29.29&52.01&50.99&\textbf{53.74}\\ \hline
\textbf{Average MCC} &47.58&45.03&41.82&42.14&42.96&46.71&44.73&41.86&30.73&52.36&52.18&\textbf{53.98}\\ \hline
\textbf{Average AUC-PR} &52.42&51.51&51.56&56.43&51.85&53.85&51.98&55.63&36.1&58&57.10&\textbf{58.89}\\ \hline
\end{tabular}}
\end{table*}

\begin{table}[]
\centering
\caption{Average performance comparison in terms of F1-score, MCC, and AUC-PR of the proposed ISFFSVM against the baseline SFFSVM on the Schizophrenia
dataset.}
\label{tab:Schezophrenia-Results-table}
\resizebox{8cm}{!}{%
\begin{tabular}{|l|c|c|c|}
\hline
Model  & F1-score & MCC & AUC-PR \\ \hline
SFFSVM \cite{ren2023slack}  & 62.73 & 23.16 & 62.76 \\  \hline
ISFFSVM (Proposed) & 63.67 & 24.85 & 62.92  \\ \hline
\end{tabular}%
}
\end{table}

\subsection{Evaluation on Schizophrenia Dataset}
To further demonstrate the effectiveness of the proposed ISFFSVM model, we conducted an evaluation using a dataset for diagnosing Schizophrenia patients. This dataset was sourced from the Center for Biomedical Research Excellence (COBRE) (available at \url{http://fcon_1000.projects.nitrc.org/indi/retro/cobre.html}). It consists of 72 subjects diagnosed with Schizophrenia, aged between 18 and 65 years (mean age of 38.1 ± 13.9 years), and 74 healthy control subjects, also aged 18 to 65 years (mean age of 35.8 ± 11.5 years). The features were extracted following the procedure described in \cite{tanveer2022intuitionistic}. Table \ref{tab:Schezophrenia-Results-table} demonstrates the performance of the proposed ISFFSVM model against the baseline SFFSVM, focusing on three key metrics: F1-score, MCC, and AUC. The proposed ISFFSVM achieved an F1-score of $63.67$, compared to $62.73$ obtained by the baseline SFFSVM. The improvement of approximately 0.94 in the F1-score demonstrates that ISFFSVM provides better precision and recall balance, particularly in managing class imbalance within the Schizophrenia dataset. Additionally, the proposed ISFFSVM model achieved an MCC of $24.85$, compared to $23.16$ for the SFFSVM. This increase of $1.69$ suggests that ISFFSVM delivers a more robust performance. Furthermore, the ISFFSVM model achieved an AUC of $62.92$, slightly higher than the AUC of $62.76$ obtained by the baseline SFFSVM. Overall, the proposed ISFFSVM model demonstrates superior performance compared to the baseline SFFSVM model across all three evaluation metrics: F1-score, MCC, and AUC. These results indicate that the proposed ISFFSVM model offers a more balanced and robust approach to handling the class imbalance in the Schizophrenia dataset, confirming its effectiveness in real-world medical diagnostic applications.

\subsection{Sensitivity Analysis}
\begin{figure}[htp]
    \centering
    \includegraphics[width=1\linewidth]{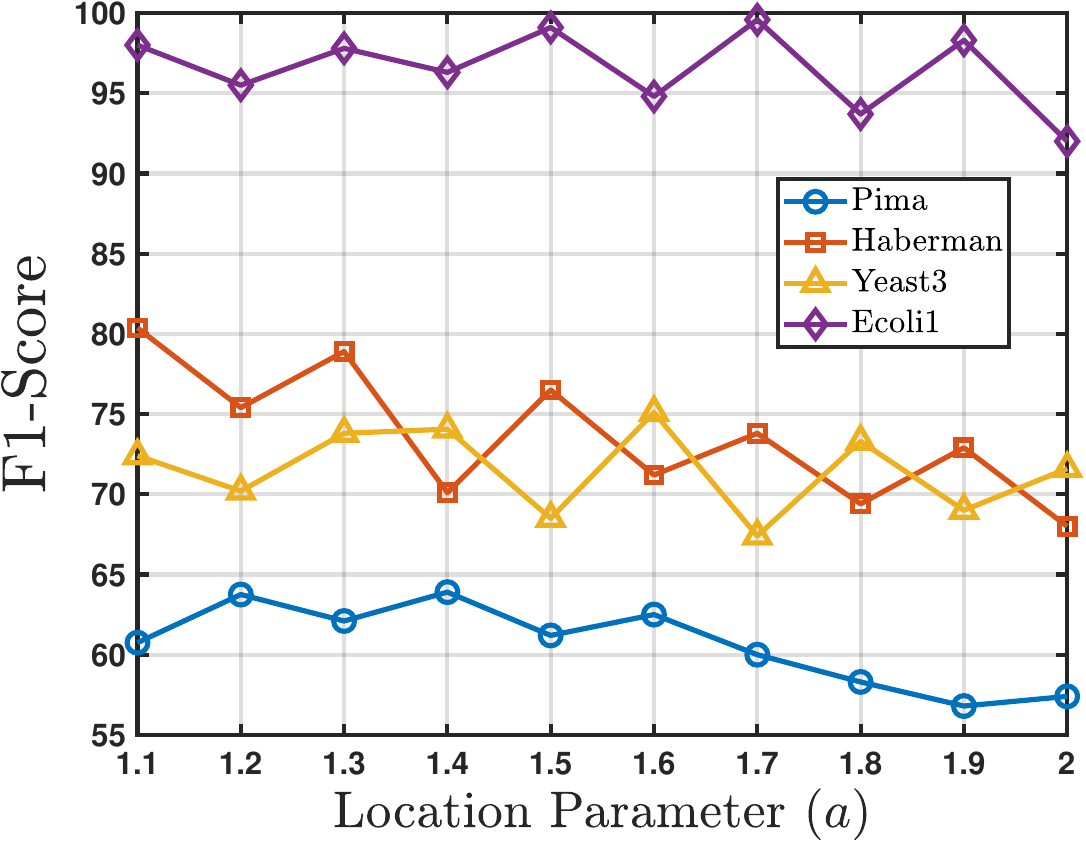}
    \caption{Sensitivity analysis of the location parameter \(a\) for the proposed ISFFSVM model, showing the F1-score values corresponding to different values of \(a\) (ranging from 1.1 to 2) across four datasets: Pima, Haberman, Yeast3, and Ecoli1.}
    \label{fig:sensitivity-analysis}
\end{figure}

To show the sensitivity of the location parameter \(a\), we plotted the F1-score corresponding to each possible value of \(a\) in the range \([1.1: 0.1: 2]\) for four datasets: Pima, Haberman, Yeast3, and Ecoli1 (see Fig. \ref{fig:sensitivity-analysis}). The analysis reveals that the F1-score is highly dependent on the value of \(a\), with the best scores achieved at specific values for each dataset: $63.75$ at \(a = 1.2\) for Pima, $80.37$ at \(a = 1.1\) for Haberman, $75.1$ at \(a = 1.6\) for Yeast3, and $99.58$ at \(a = 1.7\) for Ecoli1. For the Pima dataset, the F1-score shows moderate fluctuations around the optimal value, indicating a moderate sensitivity to \(a\). The Haberman dataset demonstrates a steep decline in F1-score when deviating from \(a = 1.1\), reflecting a high sensitivity to location parameter. In contrast, the Yeast3 dataset shows notable, though less pronounced, variability in F1-scores, highlighting the impact of \(a\) on performance. The Ecoli1 dataset displays relatively stable F1-scores with minimal variation, except for a few significant changes, suggesting lower sensitivity to \(a\) in this context. Overall, this analysis underscores the critical role of carefully tuning the location parameter \(a\) to maximize model performance across different datasets, as even small changes in \(a\) can lead to substantial variations in F1-scores, particularly in datasets with greater class imbalance or variability.

\section{Conclusions and Future Scope}
In this study, we have enhanced the existing slack-factor-based fuzzy support vector machine  (SFFSVM) by introducing a novel location parameter, resulting in the improved slack-factor-based fuzzy support vector machine  (ISFFSVM). In the SFFSVM model, misclassified majority class samples with slack factor values less than $2$ are assigned a uniform membership value. However, the value of $2$ is not always optimal, as it may lead to the misclassification of correctly classified minority samples when shifting the DEC hyperplane (\(M_{dec}\)). The incorporation of the location parameter addresses this issue by optimizing the decision boundary adjustment, preventing the DEC hyperplane from extending beyond a specific threshold and thereby mitigating the risk of misclassifying minority class samples. Extensive experimentation on diverse real-world KEEL datasets demonstrates that the proposed ISFFSVM consistently outperforms baseline models in terms of F1-score, MCC, and AUC-PR.
\par
Despite these improvements, the proposed ISFFSVM model, being a variant of SVM, faces challenges in handling large-scale imbalanced datasets due to its inherent computational complexity. Future research can focus on integrating the ISFFSVM model with algorithms designed to efficiently manage large-scale data. Additionally, researchers could focus on developing adaptive methods to dynamically and efficiently adjust the location parameter $a$ during the training process, thereby eliminating the need for manual tuning. Such methods could leverage meta-heuristic optimization algorithms or self-adaptive mechanisms to automate the adjustment process.

\section*{Acknowledgment}
This work is supported by the Science and Engineering Research Board (SERB), Government of India, through the Mathematical Research Impact-Centric Support (MATRICS) scheme under grant MTR/2021/000787. Mushir Akhtar acknowledges the Council of Scientific and Industrial Research (CSIR), New Delhi, for providing fellowship for his research under grant number 09/1022(13849)/2022-EMR-I.

\bibliographystyle{IEEEtranN}
\bibliography{refs}
\end{document}


\title{Supplementary material for the manuscript ``Enhancing Imbalance Learning: A Novel Slack-Factor Fuzzy SVM Approach"}
\author{M. Tanveer{$^*$}, \IEEEmembership{Senior Member,~IEEE}, Anushka Tiwari, Mushir Akhtar, \IEEEmembership{Graduate Student Member,~IEEE}, C.T. Lin, \IEEEmembership{Fellow,~IEEE}  
\thanks{ \noindent $^*$Corresponding Author\\
    M. Tanveer, Anushka Tiwari, and Mushir Akhtar are with the Department of Mathematics, Indian Institute of Technology Indore, Simrol, Indore, 453552, India (e-mail: mtanveer@iiti.ac.in, anushkatiwari9911@gmail.com, and phd2101241004@iiti.ac.in). \\
    Chin-Teng Lin is with School of Computer Science, Human-Centric AI Centre, University of Technology Sydney, Australia (e-mail: Chin-Teng.Lin@uts.edu.au).
    }}

\maketitle

\section{Numerical Experiments}
The comprehensive results for the low $IR$ datasets, evaluated using the metrics F1-score, MCC, and AUC-PR, are provided in Tables \ref{Table1}, \ref{Table2}, and \ref{Table3}, respectively. Similarly, the detailed results for high $IR$ datasets, in terms of F1-score, MCC, and AUC-PR, are shown in Tables \ref{Table4}, \ref{Table5}, and \ref{Table6}.

\subsection{Statistical Analysis}
To further validate the effectiveness of the proposed ISFFSVM model, we performed a comprehensive statistical analysis using ranking tests, the Friedman test, and the Nemenyi post hoc test. Since the F1-score is a critical measure for evaluating performance in class imbalance scenarios, the statistical tests are conducted based on the F1-score results.\\
\textbf{Ranking test}: To provide a more precise evaluation of the models' performance, it is essential to rank them individually on each dataset rather than depending solely on the average F1-score. This approach prevents instances where excellent performance on a few datasets might compensate for poor performance on others, thereby offering a more balanced assessment. In the ranking scheme, each model is assigned a rank for every dataset, where models with poorer performance receive a higher rank, while those with better performance are given a lower rank. Let $D$ number of datasets be used to evaluate $l$ models, and let $\Delta^d_{m}$ represent the rank of the $m^{th}$ model on the $d^{th}$ dataset. The $m^{th}$ algorithm's average rank is then calculated as follows: 
 \begin{equation}
    R_{m}=\frac{\sum_{d=1}^D\Delta_m^d}{D}.
\end{equation}
Table \ref{Rank-Table} shows the average ranks of the proposed ISFFSVM model and the baseline models for both low $IR$ and high $IR$ datasets. The results indicate that the proposed ISFFSVM model achieves a significantly lower average rank compared to the competing models. Specifically, ISFFSVM obtains an average rank of $3.07$ on the low $IR$ datasets, demonstrating its superior performance in handling moderately imbalanced data. Similarly, on the high $IR$ datasets, the ISFFSVM model achieves an even lower average rank of $2.56$, highlighting its robustness and effectiveness in managing severe class imbalance scenarios. These lower rank values indicate that the proposed model consistently outperforms the baseline models across a wide range of datasets, reinforcing its superiority in handling class imbalance.\\
\begin{table*}
\caption{Presents the average ranks of the proposed ISFFSVM model and baseline models on datasets with high $IR$.}
\label{Rank-Table}
\resizebox{\textwidth}{!}{
\tiny
\begin{tabular}{lcccccccccccc}
\hline
&DEC \cite{veropoulos1999controlling}&FSVMCIL$_{exp}$ \cite{batuwita2010fsvm}&FSVMCIL$_{lin}$
\cite{batuwita2010fsvm}&CKAFSVM \cite{wang2020centered}&MWMOTE \cite{barua2012mwmote}&PFSMOTE \cite{gazzah2008new}&SMOTETL \cite{batista2004study}&HUE \cite{ng2020hashing}&CNB \cite{rennie2003tackling}&SFFSVM \cite{ren2023slack}&SMOTE-SVM \cite{guo2024adaptive}&ISFFSVM\\
\hline
\textbf{Average Rank on Low $IR$} &4.48&7.78&8.81&6.81&6.22&5.30&6.96&9.33&8.67&4.52&5.15&\textbf{3.07}\\ 
\hline
\textbf{Average Rank on High $IR$} &5.94&6.13&6.16&7.44&7.09&6.59&7.25&7.66&10.56&4.91&5.72&\textbf{2.56}\\ \hline
\end{tabular}}
\end{table*}
\textbf{Friedman Test} \cite{demvsar2006statistical}: The Friedman test is used to evaluate the significance of differences between various models. In this test, each model is independently ranked across all datasets. The null hypothesis assumes that there is no significant difference among the models, meaning that the average ranks of all models are expected to be equal. The Friedman test follows the chi-squared distribution ($\chi^2_{F}$) with $l-1$ degrees of freedom (d.o.f.), where $l$ is the number of models being compared. Now,
\begin{align}
    \chi_F^2=& \frac{12D}{l(l+1)}\left(\sum_{m=1}^l R_m^2-\frac{l(l+1)^2}{4}\right), \\
    F_F=&\frac{(D-1)\chi_F^2}{D(l-1)-\chi_F^2}, 
\end{align}\\
where the distribution of $F_{F}$ has $(l-1)$ and $(l-1)(D-1)$ d.o.f.. The results of the Friedman test for both low $IR$ and high $IR$ datasets are presented in Table \ref{tab:Friedman test-table}. For low $IR$ datasets ($l=12$ and $D=15$), we get $\chi^2_{F}$=$36.2384$ and $F_{F}$=$3.9401$ and for high $IR$ datasets ($l=12$ and $D=33$), we get $\chi^2_{F}$=$98.9688$ and $F_{F}$=$11.9948$ . According to the statistical F-distribution table, $F_{F}$(11, 154)=$1.85$ and  $F_{F}$(11, 352)=$1.80$ for low and high $IR$ datasets, respectively at the $5$\% level of significance. We reject the null hypothesis due to the fact that $3.940>1.85$ and $11.9948>1.80$. As a result, the models differ significantly on high $IR$ datasets as well as on low $IR$ datasets. Now, we examine whether there is a significant difference between the models using the Nemenyi post hoc test.\\
\begin{table}[]
\centering
\caption{Illustrate the Friedman test results for both low $IR$ and high $IR$ datasets, demonstrating the statistical significance of the performance differences among the proposed ISFFSVM model and the baseline models. }
\label{tab:Friedman test-table}
\resizebox{9cm}{!}{%
\begin{tabular}{|l|c|c|c|c|c|c|}
\hline
Dataset              & $l$ & $D$  & $\chi^2_F$& $F_F$     & $F((l-1),(l-1)(D-1))$ & \begin{tabular}[c]{@{}c@{}}Significant difference\\ (As per Friedman test)\end{tabular}  \\ \hline
Low $IR$ dataset & 12 & 15 & 36.2384   & 3.9401  &    1.85      & Yes                    \\ \hline
High $IR$ dataset & 12 & 33 & 98.9688  & 11.9948 &    1.8    & Yes                    \\ \hline
\end{tabular}%
}
\end{table}
\textbf{Nemenyi post hoc test} \cite{demvsar2006statistical}: It compares the performance of each pair of models to determine whether their average ranks differ significantly. The Nemenyi test is based on the premise that if the performance ranks of two models differ by more than a critical difference ($C.D.$), then their performance is statistically different. The $C.D.$ is given by 
\begin{align}
    C.D.=q_{\alpha}\left(\sqrt{\frac{l(l+1)}{6D}}\right),
\end{align}
where $q_{\alpha} \left(\alpha=0.05\right)$ is the critical value for the two-tailed Nemenyi test from the distribution table. After calculation, we get $C.D.$ as $4.30$ and $2.9$ for low and high $IR$ datasets, respectively. If the difference between average ranks of the two models is more than $C.D.$, then the models are considered to be significantly different. Tables \ref{tab:Nemenyi-table-low} and \ref{tab:Nemenyi-table-high} present the results of the Nemenyi post hoc test for the low $IR$ and high $IR$ datasets, respectively. 

\begin{table}[]
\centering
\caption{Nemenyi post hoc test results showing the differences in rankings between the proposed ISFFSVM model and baseline models on the low $IR$ dataset.}
\label{tab:Nemenyi-table-low}
\resizebox{9cm}{!}{%
\begin{tabular}{|l|c|c|c|}
\hline
Model & Average rank & Rank difference & \begin{tabular}[c]{@{}c@{}}Significant difference\\ (As per Nemenyi test)\end{tabular} \\ \hline
DEC \cite{veropoulos1999controlling} & 4.48 & 1.41 & No \\ \hline
FSVMCIL$_{exp}$ \cite{batuwita2010fsvm} & 7.78 & 4.71 & Yes \\ \hline
FSVMCIL$_{lin}$ \cite{batuwita2010fsvm} & 8.81 & 5.74 & Yes \\ \hline
CKAFSV \cite{wang2020centered} & 6.81 & 3.74 & No \\ \hline
MWMOTE \cite{barua2012mwmote} & 6.22 & 3.15 & No \\ \hline
PFSMOTE \cite{gazzah2008new} & 5.3 & 2.23 & No \\ \hline
SMOTETL \cite{batista2004study} & 6.96 & 3.89 & No \\ \hline
HUE \cite{ng2020hashing} & 9.33 & 6.26 & Yes \\ \hline
CNB \cite{rennie2003tackling} & 8.67 & 5.6 & Yes \\ \hline
SFFSVM \cite{ren2023slack} & 4.52 & 1.45 & No \\ \hline
SMOTE-SVM \cite{guo2024adaptive} & 5.15 & 2.08 & No \\ \hline
ISFFSVM & 3.07 & - &  N/A \\ \hline
\end{tabular}%
}
\end{table}

\begin{table}[htp]
\centering
\caption{Nemenyi post hoc test results showing the differences in rankings between the proposed ISFFSVM model and baseline models on the high $IR$ dataset.}
\label{tab:Nemenyi-table-high}
\resizebox{9cm}{!}{%
\begin{tabular}{|l|c|c|c|}
\hline
Model & Average rank & Rank difference & \begin{tabular}[c]{@{}c@{}}Significant difference\\ (As per Nemenyi test)\end{tabular} \\ \hline
DEC \cite{veropoulos1999controlling} & 5.94 & 3.38 & Yes \\ \hline
FSVMCIL$_{exp}$ \cite{batuwita2010fsvm} & 6.13 & 3.57 & Yes \\ \hline
FSVMCIL$_{lin}$ \cite{batuwita2010fsvm} & 6.16 & 3.60 & Yes \\ \hline
CKAFSV \cite{wang2020centered} & 7.44 & 4.88 & Yes \\ \hline
MWMOTE \cite{barua2012mwmote} & 7.09 & 4.53 & Yes \\ \hline
PFSMOTE \cite{gazzah2008new} & 6.59 & 4.03 & Yes \\ \hline
SMOTETL \cite{batista2004study} & 7.25 & 4.69 & Yes \\ \hline
HUE \cite{ng2020hashing} & 7.66 & 5.1 & Yes \\ \hline
CNB \cite{rennie2003tackling} & 10.56 & 8 & Yes \\ \hline
SFFSVM \cite{ren2023slack} & 4.91 & 2.35 & No \\ \hline
SMOTE-SVM \cite{guo2024adaptive} & 5.72 & 3.16 & Yes \\ \hline
ISFFSVM & 2.56 & - &  N/A \\ \hline
\end{tabular}%
}
\end{table}
The average rank differences on the high $IR$ datasets between the proposed ISFFSVM model and  majority of the baseline models show that the proposed ISFFSVM model differs significantly from the existing baseline models. On low $IR$ datasets, its average rank does not significantly differ from the baselinemodels;, however it is clear from Table \ref{Rank-Table} that it outperforms the baseline models in terms of average ranks.


\begin{table*}
\centering
\caption{Comparative analysis of F1-scores for the proposed ISFFSVM model and various baseline models on datasets with low $IR$.}
\label{Table1}
 \resizebox{1\textwidth}{!}{
 \tiny
\begin{tabular}{lcccccccccccc}
\hline
&DEC \cite{veropoulos1999controlling}&FSVMCIL$_{exp}$ \cite{batuwita2010fsvm}&FSVMCIL$_{lin}$
\cite{batuwita2010fsvm}&CKAFSVM \cite{wang2020centered}&MWMOTE \cite{barua2012mwmote}&PFSMOTE \cite{gazzah2008new}&SMOTETL \cite{batista2004study}&HUE \cite{ng2020hashing}&CNB \cite{rennie2003tackling}&SFFSVM \cite{ren2023slack}&SMOTE-SVM \cite{guo2024adaptive}&ISFFSVM\\
\hline
Pima&65.1$\pm$3.86&66.81$\pm$3.75&66.03$\pm$3.33&62.39$\pm$4.84&62.21$\pm$5.41&62.43$\pm$4.25&58.09$\pm$5.77&64.79$\pm$3.78&53.66$\pm$4.4&61.93$\pm$12.29&57.71$\pm$6.57&63.75$\pm$4.43\\
Haberman&38.51$\pm$13.39&44.64$\pm$7.67&45.85$\pm$7.15&30.26$\pm$10.88&36.74$\pm$10.62&42.89$\pm$8.73&32.89$\pm$8.77&41.69$\pm$8.6&46.78$\pm$8.53&80.91$\pm$3.49&80.00$\pm$3.5&80.37$\pm$4.55\\
Vehicle0&94.58$\pm$3.08&93.07$\pm$3.38&92.59$\pm$3.34&93.73$\pm$2.94&94.11$\pm$2.79&94.35$\pm$2.45&94.66$\pm$2.52&89.95$\pm$2.64&53.87$\pm$3.97&94.69$\pm$2.46&92.00$\pm$5.23&95.03$\pm$9.57\\
Spect&50.59$\pm$10.88&56.54$\pm$12.19&54.76$\pm$11.24&52.45$\pm$11.33&49.59$\pm$12.12&46.83$\pm$12.14&51.29$\pm$10.11&51.54$\pm$9.81&22.09$\pm$9.49&57.24$\pm$13.26&61.76$\pm$7.65&60.98$\pm$12.5\\
Yeast1&60.57$\pm$4.96&57.8$\pm$4.14&56.68$\pm$3.72&60.38$\pm$5.75&51.93$\pm$5.51&55.18$\pm$6.41&48.02$\pm$5.48&55.02$\pm$3.57&50.78$\pm$4.8&55.21$\pm$9.57&51.00$\pm$9.12&56.24$\pm$9.09\\
Newthyroid1&94.73$\pm$5.95&94.76$\pm$6.1&97.27$\pm$4.11&91.99$\pm$8.32&94.74$\pm$7.28&94.43$\pm$5.75&96.41$\pm$4.24&93.13$\pm$6.45&90.5$\pm$6.3&94.8$\pm$5.71&96.0$\pm$5.1&95.86$\pm$4.73\\
Newthyroid2&95$\pm$5.54&94.35$\pm$8.06&94.08$\pm$7.04&91.56$\pm$7.51&95.17$\pm$5.39&94.4$\pm$7.5&96.25$\pm$5.15&92.86$\pm$6.5&89.47$\pm$6.49&93.07$\pm$5.59&94.21$\pm$5.55&94.44$\pm$5.65\\
Segment0&98.74$\pm$0.98&98.84$\pm$0.93&98.79$\pm$0.88&98.46$\pm$1.09&98.92$\pm$0.87&98.78$\pm$0.97&98.7$\pm$1.17&96.87$\pm$1.65&38.98$\pm$1.5&98.92$\pm$0.74&94.56$\pm$1.3&98.88$\pm$0.8\\
Glass6&79.04$\pm$12.2&73.78$\pm$11.57&68.44$\pm$10.7&78.76$\pm$13.07&81.31$\pm$11.75&82.5$\pm$10.41&81.93$\pm$10.03&78.41$\pm$9.31&72.46$\pm$8.87&83.56$\pm$10.17&84.21$\pm$11.23&86.67$\pm$13.25\\
Yeast3&74.85$\pm$4.16&71.7$\pm$4.87&71.03$\pm$4.69&78.03$\pm$3.74&68.28$\pm$9.48&71.23$\pm$9.94&69.22$\pm$10.15&72.01$\pm$3.77&61.01$\pm$4.34&76.74$\pm$4.37&69.81$\pm$5.6&76.1$\pm$6.1\\
Pageblocks0&83.25$\pm$2.17&79.19$\pm$2.21&80.22$\pm$2.4&83.89$\pm$2.26&84.83$\pm$2.66&86.51$\pm$1.97&82.75$\pm$2.07&80.73$\pm$2.05&61.77$\pm$4.03&83.93$\pm$1.99&86.45$\pm$2.01&84.5$\pm$2.05\\
Yeast05679vs4&45.2$\pm$9.48&42.23$\pm$7.78&43.26$\pm$8.34&49.79$\pm$14.92&40.42$\pm$12.64&39.27$\pm$13.36&41.34$\pm$13.36&42.97$\pm$5.78&43$\pm$9.61&45.13$\pm$9.95&45.66$\pm$9.01&47.33$\pm$9.93\\
Liver&60.23$\pm$6.56&61.03$\pm$8.73&62.12$\pm$6.72&55.77$\pm$8.17&60.64$\pm$7.3&58.72$\pm$6.74&59.12$\pm$7.65&60.95$\pm$5.88&61.04$\pm$5.06&62.65$\pm$5.94&60.78$\pm$6.01&62.76$\pm$6.94\\
Iris&91.78$\pm$5.86&92.98$\pm$5.44&92.41$\pm$5.71&92.24$\pm$4.85&92.6$\pm$4.84&91.75$\pm$5.25&91.97$\pm$5.7&93.22$\pm$4.64&64.78$\pm$4.97&99.01$\pm$2.94&93.34$\pm$1.5&99.61$\pm$1.3\\
Ecoli1&75.56$\pm$6.39&76.31$\pm$5.36&76.56$\pm$5.88&75.6$\pm$6.57&76.45$\pm$7.56&82.1$\pm$6.24&74.37$\pm$8.49&78.23$\pm$5.33&66.04$\pm$5.65&99.13$\pm$1.51&97.65$\pm$1.4&99.62$\pm$1.11\\
\hline
\textbf{Average}&73.85&73.6&73.34&73.02&72.53&73.43&71.8&72.83&58.42&79.13&77.68&\textbf{80.14}\\
\hline
\end{tabular}}
\end{table*} 

\begin{table*}
\centering
\caption{Comparative analysis of the MCC scores of the proposed ISFFSVM model against several baseline models on low $IR$ datasets.}
\label{Table2}
 \resizebox{1\textwidth}{!}{
\begin{tabular}{lcccccccccccc}
\hline
&DEC \cite{veropoulos1999controlling}&FSVMCIL$_{exp}$ \cite{batuwita2010fsvm}&FSVMCIL$_{lin}$
\cite{batuwita2010fsvm}&CKAFSVM \cite{wang2020centered}&MWMOTE \cite{barua2012mwmote}&PFSMOTE \cite{gazzah2008new}&SMOTETL \cite{batista2004study}&HUE \cite{ng2020hashing}&CNB \cite{rennie2003tackling}&SFFSVM \cite{ren2023slack}&SMOTE-SVM \cite{guo2024adaptive} &ISFFSVM\\
\hline
Pima&44.11$\pm$6.24&46.32$\pm$6.31&45.28$\pm$5.61&41.36$\pm$7.78&40.53$\pm$8.26&43.18$\pm$5.97&35.46$\pm$8.2&44.42$\pm$5.62&25.77$\pm$7.02&45.92$\pm$9.67&45.97$\pm$7.98&46.09$\pm$6.81\\
Haberman&24.46$\pm$12.41&23.18$\pm$10.35&23.41$\pm$10.25&18.44$\pm$12.36&13.19$\pm$10.92&26.48$\pm$11.26&9.18$\pm$7.71&17.16$\pm$11.45&27.27$\pm$11.44&12.21$\pm$11.43&10.28$\pm$10.97&13.34$\pm$11.32\\
Vehicle0&93$\pm$3.97&90.99$\pm$4.41&90.37$\pm$4.35&91.94$\pm$3.75&92.37$\pm$3.59&92.73$\pm$3.16&93.14$\pm$3.23&87.04$\pm$3.43&37.63$\pm$6.63&93.12$\pm$3.12&96.63$\pm$2.5&93.57$\pm$2.5\\
Spect&40.88$\pm$13.16&47.71$\pm$14.21&43.7$\pm$13.59&42.15$\pm$13.78&37.42$\pm$14.62&35.78$\pm$13.16&40.1$\pm$11.56&37.82$\pm$13.66&3.08$\pm$6.11&36.72$\pm$12.63&35.94$\pm$8.78&39.95$\pm$15.24\\
Yeast1&53$\pm$5.83&49.03$\pm$5.21&47.77$\pm$4.77&54.08$\pm$6.63&41.47$\pm$6.95&46.49$\pm$7.78&37.19$\pm$6.65&45.89$\pm$4.76&40.28$\pm$6.45&38.31$\pm$6.01&38.48$\pm$6.98&38.97$\pm$7.57\\
Newthyroid1&93.9$\pm$6.88&94$\pm$6.88&96.84$\pm$4.74&91.08$\pm$8.84&94.08$\pm$8.02&93.62$\pm$6.6&95.83$\pm$4.95&92.12$\pm$7.35&88.94$\pm$7.37&94.12$\pm$6.35&95.16$\pm$5.96&95.27$\pm$5.3\\
Newthyroid2&94.31$\pm$6.05&93.54$\pm$9.16&93.21$\pm$7.95&90.55$\pm$8.19&94.42$\pm$6.21&93.8$\pm$8.13&95.64$\pm$5.97&91.84$\pm$7.29&87.82$\pm$7.75&92.77$\pm$6.19&93.00$\pm$5.86&95.31$\pm$5.49\\
Segment0&98.53$\pm$1.13&98.66$\pm$1.08&98.6$\pm$1.02&98.21$\pm$1.26&98.75$\pm$1.01&98.59$\pm$1.11&98.5$\pm$1.34&96.37$\pm$1.92&31.39$\pm$2.61&98.75$\pm$0.86&91.20$\pm$0.87&98.7$\pm$0.93\\
Glass6&78.32$\pm$11.37&71.86$\pm$12.35&66.46$\pm$11.16&79.01$\pm$11.63&80.75$\pm$11.48&81.94$\pm$10.02&81.17$\pm$9.95&75.85$\pm$10.65&69.09$\pm$10.2&81.9$\pm$11.27&82.96$\pm$12.13&85.45$\pm$14.41\\
Yeast3&72.11$\pm$4.64&69.33$\pm$5.38&68.89$\pm$4.92&75.43$\pm$4.21&64.98$\pm$10.51&68.42$\pm$10.39&66.08$\pm$11.04&70.17$\pm$3.98&58.59$\pm$4.68&74.06$\pm$4.93&76.81$\pm$5.50&73.41$\pm$6.51\\
Pageblocks0&81.53$\pm$2.36&77.85$\pm$2.19&78.78$\pm$2.45&82.28$\pm$2.47&83.31$\pm$2.92&85$\pm$2.2&80.91$\pm$2.3&79.49$\pm$2.17&57.45$\pm$4.54&82.31$\pm$2.2&84.85$\pm$2.18&82.9$\pm$2.27\\
Yeast05679vs4&40.06$\pm$11.26&37.63$\pm$9.59&39.14$\pm$9.66&45.96$\pm$15.63&34.49$\pm$13.42&33.28$\pm$14.55&35.42$\pm$15.02&40.08$\pm$7.45&38.25$\pm$11.8&39.91$\pm$11.3&44.73$\pm$10.76&42.41$\pm$11.51\\
Liver&31.96$\pm$10.52&32.23$\pm$13.79&34.68$\pm$11&26.81$\pm$12.18&32.34$\pm$11.48&28.25$\pm$10.45&30.82$\pm$11.22&33.1$\pm$9.95&22.8$\pm$11.65&40.02$\pm$8.83&41.67$\pm$9.34&40.86$\pm$11.01\\
Iris&87.93$\pm$8.71&89.69$\pm$7.97&88.88$\pm$8.43&88.67$\pm$7.06&89.27$\pm$7.05&87.8$\pm$7.82&88.3$\pm$8.4&90.19$\pm$6.7&45.27$\pm$9.01&98.54$\pm$4.32&98.95$\pm$0.92&99.43$\pm$1.93\\
Ecoli1&67.93$\pm$8.61&69.24$\pm$7.21&69.6$\pm$7.96&68.07$\pm$8.75&69.25$\pm$9.95&76.82$\pm$8.09&66.58$\pm$11.07&71.8$\pm$7.09&55.94$\pm$7.87&99.05$\pm$1.65&98.47$\pm$0.87&99.58$\pm$1.21\\
\hline
\textbf{Average}&66.8&66.08&65.71&66.27&64.44&66.15&63.62&64.89&45.97&68.52&69.01&\textbf{69.68}\\\hline
\end{tabular}}
\end{table*}


\begin{table*}
\centering
\caption{Comparative analysis of AUC-PR scores of the proposed ISFFSVM model against several baseline models on low $IR$ datasets.}
\label{Table3}
 \resizebox{1\textwidth}{!}{
\begin{tabular}{lcccccccccccc}
\hline
&DEC \cite{veropoulos1999controlling}&FSVMCIL$_{exp}$ \cite{batuwita2010fsvm}&FSVMCIL$_{lin}$ \cite{batuwita2010fsvm}&CKAFSVM \cite{wang2020centered}&MWMOTE \cite{barua2012mwmote}&PFSMOTE \cite{gazzah2008new}&SMOTETL \cite{batista2004study}&HUE \cite{ng2020hashing}&CNB \cite{rennie2003tackling}&SFFSVM \cite{ren2023slack}&SMOTE-SVM\cite{guo2024adaptive}&ISFFSVM\\
\hline
Pima&65.56$\pm$5.78&66.67$\pm$6.42&64.19$\pm$5.36&67.71$\pm$5.77&61.89$\pm$6.97&66.58$\pm$5.77&59.3$\pm$6.36&67.45$\pm$4.79&48.97$\pm$6.25&71.24$\pm$5.94&69.04$\pm$5.7&71.69$\pm$5.5\\
Haberman&43.77$\pm$12.42&44.07$\pm$9.16&44.84$\pm$9.25&39.18$\pm$10.56&34.84$\pm$7.9&42.99$\pm$8.44&30.51$\pm$6.31&39.19$\pm$7.8&45.42$\pm$10.77&76.62$\pm$6.96&75.74$\pm$4.89&77.5$\pm$5.67\\
Vehicle0&99.06$\pm$0.81&98.72$\pm$1.12&98.63$\pm$1.33&98.89$\pm$0.85&98.88$\pm$0.91&99.1$\pm$0.66&99.25$\pm$0.58&95.41$\pm$2.59&59.86$\pm$7.24&99.08$\pm$0.8&95.03$\pm$0.76&99.16$\pm$0.83\\
Spect&59.35$\pm$13.42&62.07$\pm$13.71&53.64$\pm$12.45&61.24$\pm$13.15&52.72$\pm$13.35&51.89$\pm$12.36&55.19$\pm$13.65&60.33$\pm$12.25&25.22$\pm$5.59&66.87$\pm$9.88&66.48$\pm$9.45&69.08$\pm$10.64\\
Yeast1&61.9$\pm$5.87&61.7$\pm$6.7&60.39$\pm$6.64&62.47$\pm$6.73&52.86$\pm$7.86&56.86$\pm$7.51&46.88$\pm$5.91&62.76$\pm$5.28&47.87$\pm$6.55&58.34$\pm$4.64&55.90$\pm$4.78&58.07$\pm$5.21\\
Newthyroid1&98.52$\pm$3.5&99.62$\pm$1.28&99.46$\pm$1.85&98.12$\pm$3.68&98.99$\pm$2.35&99.1$\pm$1.86&99.46$\pm$1.51&98.33$\pm$3.26&98.67$\pm$2.12&99.58$\pm$1.24&96.15$\pm$0.65&99.68$\pm$1.13\\
Newthyroid2&99.21$\pm$1.69&98.71$\pm$3&98.65$\pm$2.51&98.52$\pm$2.62&98.89$\pm$1.92&99.06$\pm$2.23&99.06$\pm$2.52&98.51$\pm$2.41&97.76$\pm$3.25&99.92$\pm$1.38&98.67$\pm$0.34&99.76$\pm$0.77\\
Segment0&99.84$\pm$0.32&99.86$\pm$0.25&99.85$\pm$0.23&99.73$\pm$0.39&99.77$\pm$0.55&99.78$\pm$0.6&99.84$\pm$0.43&99.53$\pm$0.65&21.92$\pm$1.82&99.72$\pm$0.68&99.54$\pm$0.42&99.84$\pm$0.32\\
Glass6&94.52$\pm$5.26&94.13$\pm$6.33&93.74$\pm$6.08&95.3$\pm$5.14&94.22$\pm$5.38&92.61$\pm$6.33&95.02$\pm$5.44&92.22$\pm$7.58&82.01$\pm$13.84&88.9$\pm$10.41&93.70$\pm$5.87&94.49$\pm$7.14\\
Yeast3&79.85$\pm$5.75&80$\pm$6.64&81.21$\pm$6.45&80.44$\pm$6.47&71.64$\pm$10.43&76.39$\pm$10.88&72.58$\pm$12.26&81.98$\pm$4.47&51.54$\pm$7.94&81.45$\pm$5.97&82.13$\pm$4.88&81.15$\pm$5.36\\
Pageblocks0&84.96$\pm$3.62&82.44$\pm$3.18&82.42$\pm$3.67&87.32$\pm$3.11&86.33$\pm$3.62&88.87$\pm$2.46&83.63$\pm$3.17&91.76$\pm$1.97&64.41$\pm$6.31&82.89$\pm$3.79&81.15$\pm$4.78&82.01$\pm$4.49\\
Yeast05679vs4&46.49$\pm$14.65&43.29$\pm$12.61&51.19$\pm$15.53&52.3$\pm$15.04&39.95$\pm$13.99&37.44$\pm$11.7&43.14$\pm$14.64&55.88$\pm$12.38&37.81$\pm$10.36&47.06$\pm$11.8&47.61$\pm$12.87&50.52$\pm$13.39\\
Liver&67.48$\pm$7.62&67.78$\pm$7.57&68.8$\pm$7.03&61.5$\pm$7.59&66.69$\pm$7.47&64.3$\pm$7.68&65.98$\pm$6.51&66.93$\pm$6.82&58.03$\pm$9.61&67.31$\pm$6.99&67.87$\pm$5.98&69.45$\pm$7.58\\
Iris&98.12$\pm$3.59&98.74$\pm$2.26&98.3$\pm$2.38&98.15$\pm$2.09&98.27$\pm$2.53&98.57$\pm$1.75&98.21$\pm$2.19&96.77$\pm$3.82&37.06$\pm$6.96&95.13$\pm$1.94&98.06$\pm$8.67&98.37$\pm$11.47\\
Ecoli1&76.44$\pm$11.13&76.6$\pm$10.78&79.51$\pm$9.38&80.31$\pm$8.66&77.89$\pm$10.11&84.16$\pm$8.22&77.1$\pm$9.4&84.47$\pm$5.5&82.25$\pm$9&99.99$\pm$0.06&99.90$\pm$0.04&99.99$\pm$0.04\\
\hline
\textbf{Average}&78.34&78.29&78.32&78.75&75.59&77.18&75.01&79.43&57.25&82.27&81.80&\textbf{83.38}\\	\hline
\end{tabular}}
\end{table*}

\begin{table*}
\centering
 \caption{Comparative analysis of F1-scores for the proposed ISFFSVM model and various baseline models on datasets with high $IR$.}
 \label{Table4}
 \resizebox{1\textwidth}{!}{
\begin{tabular}{lcccccccccccc}
\hline
&DEC \cite{veropoulos1999controlling}&FSVMCIL$_{exp}$ \cite{batuwita2010fsvm}&FSVMCIL$_{lin}$ \cite{batuwita2010fsvm}&CKAFSVM \cite{wang2020centered}&MWMOTE \cite{barua2012mwmote}&PFSMOTE \cite{gazzah2008new}&SMOTETL \cite{batista2004study}&HUE \cite{ng2020hashing}&CNB \cite{rennie2003tackling}&SFFSVM \cite{ren2023slack}&SMOTE-SVM\cite{guo2024adaptive}&ISFFSVM\\
\hline
Abalone918&31.9$\pm$10.68&28.09$\pm$5.51&26.58$\pm$7.71&31.41$\pm$16.36&32.66$\pm$14.59&35.86$\pm$14.54&28.97$\pm$12.63&24.68$\pm$5&15.41$\pm$4.26&49.81$\pm$10.53&47.89$\pm$8.65&50.92$\pm$9.77\\
Dermatology6&62.5$\pm$21.8&71.18$\pm$20.21&63.22$\pm$21.99&38$\pm$25.21&40.99$\pm$25.32&46.36$\pm$27.96&51.33$\pm$27.89&96$\pm$5.39&91.22$\pm$8.02&96.19$\pm$7.33&85.35$\pm$6.89&97.14$\pm$21.81\\
Ecoli0137vs26&46.8$\pm$43.91&45.47$\pm$39.01&33.15$\pm$36.28&70$\pm$46.29&35.57$\pm$41.11&68.33$\pm$45.95&28.93$\pm$33.74&15.8$\pm$9.67&55.67$\pm$42&62$\pm$41.93&66.99$\pm$30.65&69.33$\pm$41.93\\
Ecoli4&81.89$\pm$12.09&81.76$\pm$13.01&67.61$\pm$14.93&86.76$\pm$10.91&79.98$\pm$12.75&80.51$\pm$12.93&78.06$\pm$14.57&56.8$\pm$11.75&48.05$\pm$9.54&77.95$\pm$13.47&77.89$\pm$12.78&78.56$\pm$16.64\\
Glass016vs2&35.51$\pm$22.74&37.87$\pm$19.09&36.07$\pm$16.94&14.53$\pm$23.44&41.26$\pm$22.71&44.42$\pm$29.5&37.1$\pm$25.36&31.91$\pm$10.22&10.53$\pm$8.35&37.55$\pm$19.5&35.87$\pm$18.65&40.28$\pm$20.2\\
Glass016vs5&67.82$\pm$27.7&36.14$\pm$21.9&43.88$\pm$25.7&53.8$\pm$33.23&60.6$\pm$32.55&58.27$\pm$30.04&55$\pm$31.62&46.36$\pm$10.85&20.61$\pm$13.13&70.28$\pm$24.72&70.67$\pm$23.86&74$\pm$27.82\\
Glass2&39.52$\pm$20.53&31.92$\pm$16.59&32.51$\pm$20.16&20.22$\pm$23.9&42.04$\pm$22.13&43.58$\pm$25.1&35.82$\pm$22.06&30.94$\pm$10.28&12.83$\pm$7.76&37.72$\pm$20.19&32.89$\pm$21.9&39.21$\pm$21\\
Glass4&72.8$\pm$24.15&61.81$\pm$21.97&66.83$\pm$20.52&69.03$\pm$26.18&72.91$\pm$22.02&81.01$\pm$20.41&87.99$\pm$15.53&55.33$\pm$11.83&36.65$\pm$11.86&69.8$\pm$22.25&69.67$\pm$19.89&73.81$\pm$20.62\\
Glass5&65.68$\pm$30.24&40.29$\pm$29.11&44.96$\pm$30.24&37$\pm$36.62&56$\pm$33.05&62.33$\pm$32.89&52.47$\pm$33.39&47.34$\pm$14.46&15.53$\pm$10.73&65.76$\pm$29.15&70.00$\pm$21.89&74.88$\pm$24.55\\
Shuttle6vs23&50.78$\pm$39.23&67.7$\pm$18.92&58.94$\pm$28.43&61.33$\pm$31.84&60$\pm$38.69&63.33$\pm$35.15&72.67$\pm$29.11&100$\pm$0&72.42$\pm$18.08&91.87$\pm$14.08&90.67$\pm$11.09&93.6$\pm$13.06\\
Shuttle-0vs4&98.6$\pm$1.54&97.58$\pm$1.69&98.32$\pm$1.51&98.42$\pm$1.84&98.43$\pm$1.73&98.32$\pm$1.57&97.88$\pm$1.8&100$\pm$0&99.26$\pm$1.17&98.91$\pm$1.25&98.89$\pm$0.6&99.1$\pm$1.4\\
WQ-R3vs5&10.1$\pm$14.56&8.35$\pm$10.75&8.26$\pm$10.35&0$\pm$0&2.8$\pm$11.26&0$\pm$0&6.39$\pm$15.44&7.64$\pm$4.07&6.4$\pm$2.47&9.88$\pm$16.18&12.09$\pm$13.65&8.92$\pm$18.39\\
WQ-R4&16.21$\pm$8.87&14.53$\pm$4.67&15.34$\pm$5.84&17.09$\pm$9.85&14.07$\pm$8.12&12.74$\pm$8.42&15.99$\pm$8.24&13.63$\pm$2.63&11.46$\pm$2.01&16.8$\pm$9.46&13.09$\pm$7.65&15.27$\pm$8.02\\
WQ-R8vs67&10.67$\pm$11.8&8.61$\pm$3.08&6.07$\pm$6.15&3.84$\pm$9.87&7.01$\pm$10.67&9.18$\pm$11.88&12.56$\pm$12.71&11.57$\pm$2.71&9.81$\pm$2.76&6.97$\pm$11.43&7.09$\pm$11.87&8.43$\pm$12.62\\
WQ-R8vs6&14.34$\pm$12.97&13.37$\pm$3.91&14.51$\pm$11.87&1.14$\pm$5.66&22.79$\pm$22.06&13.5$\pm$14.62&18.03$\pm$16.31&17.57$\pm$4.45&15.72$\pm$3.91&12.23$\pm$15.39&9.09$\pm$12.98&13.33$\pm$13.95\\
WQ-W3vs7&29.65$\pm$17.84&15.4$\pm$6.33&11.93$\pm$3.09&0$\pm$0&1.82$\pm$7.49&0$\pm$0&1.83$\pm$7.39&13.18$\pm$3.82&12.92$\pm$5.09&20.29$\pm$21.34&19.98$\pm$21.78&22.52$\pm$23.88\\
WQ-W9vs4&62$\pm$49.03&53.33$\pm$44.54&44.03$\pm$47.27&36$\pm$48.49&28.67$\pm$43.13&32$\pm$47.12&40$\pm$49.49&19.29$\pm$10.75&14.81$\pm$9.89&60.4$\pm$35.36&61.20$\pm$31.89&64.47$\pm$35.53\\
Yeast2vs8&43.39$\pm$20.02&36.15$\pm$17.88&24.55$\pm$17.74&41.01$\pm$23.96&24.92$\pm$19.74&36.47$\pm$23.33&22.49$\pm$17.5&23.89$\pm$6.09&60.58$\pm$17.3&62.91$\pm$21.31&58.00$\pm$21.76&62.98$\pm$22.64\\
Yeast1vs7&28.08$\pm$12.34&32.01$\pm$7.82&24.1$\pm$7.5&34.8$\pm$16.69&20.19$\pm$11.37&21.37$\pm$15.32&25.36$\pm$13.83&27.29$\pm$5.06&29.83$\pm$7.86&16.12$\pm$18.07&17.04$\pm$17.97&18.29$\pm$18.76\\
Yeast4&33.92$\pm$10.67&24.24$\pm$5.08&23.22$\pm$4.23&35.68$\pm$6.55&28.08$\pm$8.93&30.25$\pm$11.52&29.32$\pm$10.96&23.43$\pm$2.74&28.83$\pm$4.9&31.69$\pm$11.3&29.52$\pm$10.54&31.69$\pm$11.14\\
Yeast5&68.13$\pm$9.49&54.92$\pm$11.26&54.28$\pm$10.87&68.84$\pm$9.36&66.56$\pm$10.47&67.24$\pm$12.41&69.15$\pm$9.94&58.09$\pm$7.38&36.65$\pm$4.21&67.29$\pm$9.93&62.00$\pm$9.76&67.41$\pm$10.43\\
Aba.17vs789A&30.34$\pm$6.61&22.42$\pm$3.58&24.18$\pm$4.41&29.07$\pm$9.24&29.3$\pm$11.12&22.83$\pm$10.51&26.59$\pm$10.71&20.44$\pm$2.69&9.1$\pm$1.41&29.55$\pm$6.18&29.04$\pm$6.43&30.63$\pm$6.06\\
Aba.19vsABCD&7.35$\pm$5.49&6.02$\pm$2.2&6.38$\pm$3.26&5.87$\pm$9.2&8.7$\pm$9.4&8.88$\pm$10.43&8.79$\pm$9.26&7.05$\pm$1.52&4.36$\pm$1.94&11.21$\pm$7.17&9.77$\pm$5.64&10.21$\pm$7.4\\
Aba.21vs8&41.56$\pm$24.09&39.88$\pm$15.17&31.99$\pm$17.8&53.29$\pm$23.37&33.03$\pm$25.51&44.89$\pm$22.5&33.41$\pm$22.98&27.13$\pm$7.15&13.86$\pm$3.63&53.42$\pm$23.99&55.34$\pm$18.76&55.95$\pm$19.5\\
Aba.3vsB&98.57$\pm$4.33&99.14$\pm$3.43&96.57$\pm$6.16&99.6$\pm$2.83&98.57$\pm$4.33&100$\pm$0&100$\pm$0&85.13$\pm$12.59&50.54$\pm$9.8&96.71$\pm$6.91&96.97$\pm$4.65&97.79$\pm$5.72\\
Poker89vs5&7.04$\pm$6.97&5.89$\pm$3.61&5.3$\pm$6.19&4.32$\pm$7.96&11.37$\pm$13.26&18.44$\pm$12.26&16.31$\pm$12.21&6.65$\pm$1.97&2.1$\pm$1.01&15.76$\pm$14.98&16.7$\pm$12.76&15.91$\pm$13.75\\
Poker8vs6&63.84$\pm$29.29&27.75$\pm$16.2&48.83$\pm$32.29&42$\pm$32.2&62.6$\pm$32.56&63.8$\pm$30.7&76.4$\pm$21.07&5.02$\pm$2.36&1.68$\pm$0.96&92.4$\pm$12.55&88.9$\pm$18.54&89.93$\pm$19.14\\
Poker9vs7&20.33$\pm$35.84&37.17$\pm$26.58&7.25$\pm$9.35&0$\pm$0&24$\pm$34.36&50.67$\pm$33.16&43.33$\pm$37.04&15.05$\pm$8.84&5.83$\pm$5.41&39.04$\pm$31.02&44.99$\pm$21.98&43.98$\pm$31\\
Yeast6&42.62$\pm$10.95&39.49$\pm$6.62&35.06$\pm$6.62&49.92$\pm$8.63&41.34$\pm$12.47&39.53$\pm$13.36&38.12$\pm$11.08&19.74$\pm$2.57&27.08$\pm$3.71&42.68$\pm$12.9&39.34$\pm$11.87&42.33$\pm$12.57\\
WQ-W39vs5&11.07$\pm$11.99&11.07$\pm$5.12&9.09$\pm$9.36&3.66$\pm$8.55&7.73$\pm$13.11&2.46$\pm$7.49&3.86$\pm$7.49&7.27$\pm$1.89&4.87$\pm$1.85&9.3$\pm$12.02&10.9$\pm$11.55&11.75$\pm$12.94\\
Vowel0&99.12$\pm$1.61&98.61$\pm$2.39&98.93$\pm$2.76&99.44$\pm$2.62&98.99$\pm$3.19&99.51$\pm$1.06&99.45$\pm$1.1&83.04$\pm$5.06&57.83$\pm$5.3&99.62$\pm$1.08&93.34$\pm$1.23&99.46$\pm$1.21\\
Car&98.1$\pm$3.38&94.35$\pm$6.48&92.56$\pm$8.76&95.33$\pm$6.46&98.42$\pm$2.73&99.05$\pm$1.71&98.06$\pm$3.26&82.98$\pm$5.2&29.7$\pm$2.58&90.64$\pm$6.33&85.67$\pm$4.56&90.53$\pm$5.65\\
Shuttlec2vsc4&68.8$\pm$39.1&76.6$\pm$31.08&81$\pm$31.77&76$\pm$43.14&72.67$\pm$40.37&90$\pm$30.3&74.47$\pm$32.17&100$\pm$0&54.33$\pm$43.49&73.67$\pm$38.26&75.78$\pm$32.65&80.8$\pm$37.39\\
\hline
\textbf{Average}&47.24&43&40.47&41.74&43.15&46.82&45.03&38.8&29.29&52.01&50.99&\textbf{53.74}\\
\hline
\end{tabular}}
\end{table*}


\begin{table*}
\centering
 {\caption{Comparative analysis of the MCC scores of the proposed ISFFSVM model against several baseline models on high $IR$ datasets.}
\label{Table5}
\resizebox{1\textwidth}{!}{
\begin{tabular}{lcccccccccccc}
\hline
&DEC \cite{veropoulos1999controlling}&FSVMCIL$_{exp}$ \cite{batuwita2010fsvm}&FSVMCIL$_{lin}$ \cite{batuwita2010fsvm}&CKAFSVM \cite{wang2020centered}&MWMOTE \cite{barua2012mwmote}&PFSMOTE \cite{gazzah2008new}&SMOTETL \cite{batista2004study}&HUE \cite{ng2020hashing}&CNB \cite{rennie2003tackling}&SFFSVM \cite{ren2023slack}&SMOTE-SVM \cite{guo2024adaptive}&ISFFSVM\\
\hline
Abalone918&30.91$\pm$11.97&28.8$\pm$7.13&26.3$\pm$10.01&30.74$\pm$16.63&29.26$\pm$15.26&32.98$\pm$15.1&25.28$\pm$13.29&24.98$\pm$7.96&12.63$\pm$7.85&48.91$\pm$11.27&45.37$\pm$9.65&49.89$\pm$10.38\\ 
Dermatology6&66.64$\pm$18.79&74.25$\pm$16.97&66.24$\pm$21.35&43.19$\pm$26.67&46.26$\pm$26.1&50.94$\pm$27.85&55.2$\pm$27.7&95.96$\pm$5.44&91.28$\pm$7.84&96.31$\pm$7&95.93$\pm$4.56&97.19$\pm$5.67\\
Ecoli0137vs26&48.55$\pm$43.92&48.17$\pm$39.26&36.08$\pm$37&70$\pm$46.29&37.12$\pm$41.72&68.54$\pm$45.91&31.41$\pm$35.3&24.2$\pm$13.34&57.09$\pm$42.17&63.09$\pm$41.98&65.99$\pm$39.76&70.55$\pm$41.98\\
Ecoli4&81.5$\pm$12.63&82.07$\pm$12.77&67.7$\pm$14.74&86.73$\pm$11.02&80.13$\pm$12.39&80.55$\pm$12.98&77.78$\pm$15.04&58.24$\pm$11.88&50.22$\pm$9.75&77.87$\pm$13.79&75.47$\pm$14.76&78.5$\pm$16.97\\
Glass016vs5&68.57$\pm$27.97&38.79$\pm$21.33&46.62$\pm$25.1&54.82$\pm$33.91&60.94$\pm$33.08&59.68$\pm$30.65&55.64$\pm$32.27&50.99$\pm$9.56&19.31$\pm$15.12&35.21$\pm$20.98&37.9$\pm$19.54&37.03$\pm$21.28\\
Glass016vs2&32.92$\pm$23.52&35.53$\pm$21.09&32.73$\pm$19.03&13.79$\pm$23.19&38.28$\pm$23.33&42.79$\pm$29.88&34.31$\pm$25.67&30.92$\pm$14.27&4.45$\pm$8.38&70.91$\pm$25.07&75.9$\pm$24.76&74.53$\pm$27.99\\
Glass2&36.54$\pm$21.43&29.17$\pm$18.33&29.8$\pm$21.5&19.01$\pm$23.81&39.5$\pm$23.32&41.39$\pm$25.37&33.79$\pm$22.53&31.84$\pm$14.69&8.46$\pm$10.59&36.13$\pm$21.37&35.9$\pm$19.43&37.22$\pm$21.58\\
Glass4&73.31$\pm$23.96&63.33$\pm$21.06&67.24$\pm$20.84&70.38$\pm$25.56&74.11$\pm$21.27&81.57$\pm$19.98&88.45$\pm$14.79&56.27$\pm$11.98&35.32$\pm$14.96&69.73$\pm$22.46&70.14$\pm$19.23&73.91$\pm$20.52\\
Glass5&66.78$\pm$30.55&43.74$\pm$28.45&46.32$\pm$30.66&37.62$\pm$37.33&57.1$\pm$33.64&63.66$\pm$33.22&53.04$\pm$33.99&52.51$\pm$12.67&15.23$\pm$12.88&66.69$\pm$29.43&74.98$\pm$21.76&75.72$\pm$24.55\\
Shuttle6vs23&51.9$\pm$39.66&70.1$\pm$17.5&61.54$\pm$28.37&63.35$\pm$32.23&61.37$\pm$38.96&64.96$\pm$35.35&74.36$\pm$28.8&100$\pm$0&74.64$\pm$16.26&92.61$\pm$12.75&95.15$\pm$9.65&94.2$\pm$11.81\\
Shuttle-0vs4&98.52$\pm$1.61&97.46$\pm$1.75&98.22$\pm$1.59&98.35$\pm$1.91&98.35$\pm$1.81&98.22$\pm$1.65&97.77$\pm$1.89&100$\pm$0&99.21$\pm$1.24&98.85$\pm$1.32&99.99$\pm$0.78&99.05$\pm$1.47\\
WQ-R3vs5&10.85$\pm$15.52&9.62$\pm$12.35&8.97$\pm$12.19&0$\pm$0&2.77$\pm$11.12&0$\pm$0&6.44$\pm$15.42&12.96$\pm$8&10.55$\pm$6.34&10.37$\pm$17.8&8.9$\pm$19.23&9.53$\pm$19.25\\
WQ-R4&15.19$\pm$9.12&14.57$\pm$6.61&15.03$\pm$6.98&14.81$\pm$9.89&11.2$\pm$7.96&10.32$\pm$8.04&13.27$\pm$8.36&15.11$\pm$5.43&11.86$\pm$4.58&16.25$\pm$10.22&13.9$\pm$8.76&14.11$\pm$8.45\\
WQ-R8vs67&10.74$\pm$12.01&12.13$\pm$5.85&8.1$\pm$8.33&3.76$\pm$9.8&6.47$\pm$9.94&8.69$\pm$11.75&11.7$\pm$12.1&16.57$\pm$6.28&13.4$\pm$6.3&7.48$\pm$12.51&10.9$\pm$12.56&9.17$\pm$14.44\\
WQ-R8vs6&14.33$\pm$13.19&17.87$\pm$6.89&15.64$\pm$13.52&1.08$\pm$5.34&22.78$\pm$22.74&12.94$\pm$14.83&17.11$\pm$16.03&22.3$\pm$7.93&20.63$\pm$7.35&13.64$\pm$18.11&14.9$\pm$13.89&14.25$\pm$16.71\\
WQ-W3vs7&30.62$\pm$18.76&20.2$\pm$7.5&16.02$\pm$5.98&0$\pm$0&1.95$\pm$8.38&0$\pm$0&1.84$\pm$7.47&17.56$\pm$7.48&16.77$\pm$8.85&20.63$\pm$22.14&22.9$\pm$23.89&23.53$\pm$25.07\\
WQ-W9vs4&62$\pm$49.03&54.16$\pm$44.73&44.13$\pm$47.66&36$\pm$48.49&28.96$\pm$43.4&32$\pm$47.12&40$\pm$49.49&26.63$\pm$13.39&21.25$\pm$13.51&62.76$\pm$34.92&65.9$\pm$31.87&66.56$\pm$34.86\\
Yeast2vs8&44.49$\pm$21.08&36.83$\pm$19.29&23.31$\pm$19.17&44.38$\pm$24.79&24.13$\pm$20.18&38.54$\pm$24.58&20.86$\pm$17.28&25.2$\pm$8.24&61.3$\pm$16.98&64.91$\pm$20.38&59.47$\pm$21.44&64.62$\pm$22.16\\
Yeast1vs7&24.6$\pm$13.18&30.49$\pm$10&21.07$\pm$10.13&34.5$\pm$17.51&14.8$\pm$11.46&16.8$\pm$15.15&20.38$\pm$14.88&27.19$\pm$7.76&27.08$\pm$10.5&16.68$\pm$19.89&16.56$\pm$16.43&17.03$\pm$18.68\\
Yeast4&33.37$\pm$11.13&29.28$\pm$5.62&27.82$\pm$5.57&36.23$\pm$7.59&26.07$\pm$9.71&28.12$\pm$12.11&27.19$\pm$11.63&30.17$\pm$3.79&32.07$\pm$6.01&31.57$\pm$11.93&25.87$\pm$10.27&30.78$\pm$11.35\\
Yeast5&68.07$\pm$9.73&58.28$\pm$9.73&57.52$\pm$9.73&68.73$\pm$9.68&66.11$\pm$10.87&66.88$\pm$12.68&68.82$\pm$10.16&62.16$\pm$5.96&44.68$\pm$3.64&67.45$\pm$10.04&68.42$\pm$9.65&67.54$\pm$10.79\\
Aba.17vs789A&32.44$\pm$7.45&28.96$\pm$4.42&30.66$\pm$5.77&27.46$\pm$9.63&28.03$\pm$11.7&21.05$\pm$10.6&24.98$\pm$10.68&27.57$\pm$4.48&11.53$\pm$3.86&31.4$\pm$7.43&26.56$\pm$6.87&32.23$\pm$7.22\\
Aba.19vsABCD&7.64$\pm$5.94&7.8$\pm$4.31&7.43$\pm$5.22&5.49$\pm$8.91&8$\pm$9.13&8.27$\pm$10.31&8.01$\pm$8.84&10.03$\pm$4.36&3.78$\pm$4.57&11.4$\pm$7.83&10.73$\pm$8.76&10.86$\pm$8.22\\
Aba.21vs8&43.23$\pm$24.54&42.74$\pm$15.25&35.98$\pm$18.13&54.76$\pm$23.7&34.08$\pm$26.62&45.33$\pm$23.06&33.67$\pm$23.81&32.58$\pm$8.12&20.36$\pm$7.54&53.82$\pm$24.28&52.73$\pm$18.12&56.29$\pm$19.64\\
Aba.3vsB&98.62$\pm$4.19&99.17$\pm$3.32&96.68$\pm$5.97&99.62$\pm$2.65&98.62$\pm$4.19&100$\pm$0&100$\pm$0&86.08$\pm$11.48&56.36$\pm$8.18&96.85$\pm$6.59&99.99$\pm$4.78&97.87$\pm$5.47\\
Poker89vs5&7.95$\pm$8.23&8.91$\pm$5.95&6.56$\pm$7.51&4.21$\pm$7.79&11.12$\pm$13.2&18.2$\pm$12.6&16.06$\pm$12.23&12.59$\pm$4.96&1.25$\pm$1.83&16.4$\pm$15.21&15.5$\pm$13.89&16.61$\pm$14.05\\
Poker8vs6&66.18$\pm$28.85&35.02$\pm$15.76&53.79$\pm$29.86&45.62$\pm$33.57&65.8$\pm$31.79&66.46$\pm$30.32&78.93$\pm$18.98&11.08$\pm$5.56&1.27$\pm$1.91&93.12$\pm$11.09&90.09$\pm$13.76&90.71$\pm$18.06\\
Poker9vs7&20.75$\pm$36.43&38.61$\pm$27.24&8.3$\pm$11.47&0$\pm$0&24.99$\pm$35.63&52.78$\pm$34.22&44.99$\pm$38.12&15.64$\pm$12.45&3.65$\pm$6.22&39.52$\pm$32.02&30.99$\pm$32.87&45.15$\pm$31.95\\
Yeast6&45.05$\pm$12.11&43.57$\pm$6.85&40.95$\pm$7.22&51.45$\pm$9.1&40.55$\pm$13.08&38.82$\pm$13.8&37.34$\pm$11.68&27.29$\pm$4.2&33.89$\pm$4.87&43.29$\pm$13.35&44.86$\pm$12.44&43.05$\pm$13.64\\
WQ-W39vs5&10.92$\pm$12.06&15.37$\pm$7.88&9.74$\pm$11.1&3.58$\pm$8.44&8.04$\pm$13.99&2.46$\pm$7.54&3.57$\pm$6.95&10.69$\pm$4.58&5.49$\pm$4.74&9.6$\pm$12.75&12.9$\pm$12.56&12.76$\pm$14.29\\
Vowel0&99.05$\pm$1.75&98.5$\pm$2.56&98.86$\pm$2.89&99.43$\pm$2.65&98.94$\pm$3.22&99.47$\pm$1.15&99.41$\pm$1.2&82.34$\pm$5.11&56.26$\pm$5.85&99.59$\pm$1.17&92.73$\pm$0.9&99.42$\pm$1.31\\
Car&98.09$\pm$3.37&94.41$\pm$6.35&92.79$\pm$8.22&95.38$\pm$6.27&98.39$\pm$2.81&99.04$\pm$1.73&98.05$\pm$3.27&83.6$\pm$4.71&37.66$\pm$2.45&90.51$\pm$6.43&87.65$\pm$4.89&90.35$\pm$5.79\\
Shuttlec2vsc4&69.84$\pm$38.81&78.1$\pm$30&81.79$\pm$31.26&76$\pm$43.14&73.42$\pm$40.03&90$\pm$30.3&76.32$\pm$30.82&100$\pm$0&55.12$\pm$43.69&74.41$\pm$38&76.89$\pm$32.67&81.1$\pm$37.19\\
\hline
\textbf{Average}&$47.58$&$45.03$&$41.82$&$42.14$&$42.96$&$46.71$&$44.73$&$41.86$&$30.73$&$52.36$&$52.18$&\textbf{53.98}\\
\hline
\end{tabular}}
}
\end{table*}


\begin{table*}
\centering
 \caption{Comparative analysis of AUC-PR scores of the proposed ISFFSVM model against several baseline models on high $IR$ datasets.}
\label{Table6}
 \resizebox{1\textwidth}{!}{
\begin{tabular}{lcccccccccccc}
\hline
&DEC \cite{veropoulos1999controlling}&FSVMCIL$_{exp}$ \cite{batuwita2010fsvm}&FSVMCIL$_{lin}$ \cite{batuwita2010fsvm}&CKAFSVM \cite{wang2020centered}&MWMOTE \cite{barua2012mwmote}&PFSMOTE \cite{gazzah2008new}&SMOTETL \cite{batista2004study}&HUE \cite{ng2020hashing}&CNB \cite{rennie2003tackling}&SFFSVM \cite{ren2023slack}&SMOTE-SVM \cite{guo2024adaptive}&ISFFSVM\\
\hline
Abalone918&32.84$\pm$17.7&33.26$\pm$13.64&32.41$\pm$14.6&35.84$\pm$14.38&32.46$\pm$16.9&35.35$\pm$15.02&29.21$\pm$14&37.66$\pm$15.58&22.09$\pm$10.22&60.81$\pm$16.63&55.01$\pm$13.25&59.14$\pm$14.68\\
Dermatology6&86.4$\pm$34.06&93.83$\pm$23.29&100$\pm$0&100$\pm$0&100$\pm$0&100$\pm$0&100$\pm$0&99.66$\pm$1.35&100$\pm$0&100$\pm$0&99.93$\pm$0.2&100$\pm$0\\
Ecoli0137vs26&50.53$\pm$44.94&59.7$\pm$46.13&46.69$\pm$44.64&70.91$\pm$44.89&45.06$\pm$47.6&68.69$\pm$46.16&44.71$\pm$46.23&51$\pm$45.03&70.25$\pm$42.26&74.71$\pm$43.15&79.99$\pm$24.89&79.08$\pm$39.92\\
Ecoli4&87.79$\pm$13.49&92.37$\pm$11.16&80.68$\pm$14.67&93.58$\pm$10.17&85.53$\pm$13.02&88.71$\pm$12.13&84.99$\pm$13.12&84.46$\pm$15.79&37.2$\pm$18.46&88.39$\pm$10.42&90.99$\pm$11&90.22$\pm$12\\
Glass016vs5&82.22$\pm$19.77&76.72$\pm$24.52&75.68$\pm$25.85&75.32$\pm$26.86&77.44$\pm$22.97&77.29$\pm$25.93&77.92$\pm$26.02&91.3$\pm$16.94&13.34$\pm$4.15&34.47$\pm$21.51&37.90$\pm$21.7&37.69$\pm$21.51\\
Glass016vs2&36.11$\pm$23.19&45.97$\pm$26.61&35.91$\pm$20.52&28.56$\pm$21.26&41.23$\pm$23.83&46.69$\pm$27.17&40.06$\pm$25.46&36.13$\pm$21.21&16.57$\pm$13.9&75.71$\pm$27.28&84.90$\pm$23.89&83.5$\pm$26.03\\
Glass2&49.97$\pm$28&44.37$\pm$25.12&42.05$\pm$25.97&29.19$\pm$21.41&42.57$\pm$22.67&45.52$\pm$24.66&39.74$\pm$23.26&30.61$\pm$19.26&12.25$\pm$9.61&37.65$\pm$19.86&34.68$\pm$18.65&38.04$\pm$18.65\\
Glass4&89.23$\pm$16.74&87.51$\pm$16.22&85.25$\pm$17.86&89.71$\pm$15.42&90.28$\pm$15.33&93.33$\pm$10.57&93.51$\pm$10.67&83.55$\pm$14.56&28.38$\pm$17.76&82.22$\pm$19.16&80.00$\pm$19.87&82.41$\pm$20.92\\
Glass5&76.11$\pm$25.11&74.56$\pm$28.01&71.67$\pm$25.7&70.35$\pm$26.38&78.6$\pm$26.91&83.09$\pm$20.08&73.78$\pm$28.65&90.29$\pm$19.55&10.34$\pm$3.75&80.48$\pm$25.63&84.11$\pm$21.66&84.01$\pm$24.18\\
Shuttle6vs23&96.17$\pm$9.02&95.79$\pm$15.16&97$\pm$8.34&98.42$\pm$6.4&99.17$\pm$4.12&99.17$\pm$4.12&98.58$\pm$5.75&100$\pm$0&100$\pm$0&99.17$\pm$4.12&97.39$\pm$0.1&100$\pm$0\\
Shuttle-0vs4&99.99$\pm$0.02&99.99$\pm$0.04&99.98$\pm$0.05&99.99$\pm$0.04&100$\pm$0&99.99$\pm$0.02&99.97$\pm$0.05&100$\pm$0&100$\pm$0&100$\pm$0.01&99.99$\pm$0.2&100$\pm$0.01\\
WQ-R3vs5&9.19$\pm$12.11&9.15$\pm$12&7.31$\pm$7.92&14.76$\pm$16.82&8.21$\pm$11.91&4.91$\pm$4.69&8.33$\pm$13.32&18.87$\pm$22.32&26.13$\pm$22.89&20.89$\pm$26&20.00$\pm$25.4&21.42$\pm$26.76\\
WQ-R4&12.93$\pm$6.05&11.99$\pm$4.81&12.14$\pm$4.51&13.81$\pm$6.89&8.61$\pm$4.77&7.73$\pm$3.33&9.36$\pm$3.9&12.51$\pm$5.93&13.55$\pm$5.84&13.19$\pm$5.72&10.77$\pm$5.43&11.41$\pm$5.08\\
WQ-R8vs67&8.84$\pm$9.64&5.09$\pm$2.47&6.06$\pm$6.4&6.28$\pm$8&7.22$\pm$8.18&8.56$\pm$9.44&11.64$\pm$11.44&24.3$\pm$17.14&6.33$\pm$2.66&9.08$\pm$10.13&8.8$\pm$7.78&8.47$\pm$8.61\\
WQ-R8vs6&15.1$\pm$12.79&19.06$\pm$14.22&17.54$\pm$12.24&9.44$\pm$6.83&25.14$\pm$18.43&17.09$\pm$13.36&20.39$\pm$15.15&35.2$\pm$20.66&12.73$\pm$7.14&15.06$\pm$11.45&11.23$\pm$11.23&15.43$\pm$12.45\\
WQ-W3vs7&18.39$\pm$10.29&16.96$\pm$9.68&12.29$\pm$4.91&38.96$\pm$20.5&15.97$\pm$9.39&13.33$\pm$6.81&17.27$\pm$8.19&20.28$\pm$16.8&16.57$\pm$14.4&26.99$\pm$19.89&31.11$\pm$18.76&34.31$\pm$19.65\\
WQ-W9vs4&50.95$\pm$49.55&39.16$\pm$48.11&63.56$\pm$45.58&86.37$\pm$34.12&68.19$\pm$45.06&70.16$\pm$44.23&71.31$\pm$42.67&68.69$\pm$42.44&53.82$\pm$45.4&78.77$\pm$40.41&72.94$\pm$40&74.94$\pm$41\\
Yeast2vs8&53.87$\pm$22.11&48.82$\pm$24.3&31.91$\pm$17.52&55.17$\pm$23.01&24.5$\pm$17.09&46.69$\pm$22.67&21.49$\pm$17.55&58.26$\pm$20.01&56$\pm$18.09&59.55$\pm$20.00&63.17$\pm$21.78&63.53$\pm$21.89\\
Yeast1vs7&25.71$\pm$13.81&30.46$\pm$10.91&26.92$\pm$15.5&34.15$\pm$15.17&18.7$\pm$10.97&19.23$\pm$11.66&22.47$\pm$13.15&34.37$\pm$15.93&29.8$\pm$13.2&34.43$\pm$18.04&30.02$\pm$16.88&31.55$\pm$17.99\\
Yeast4&32.34$\pm$13.41&33.27$\pm$12.08&31.79$\pm$12.49&35.76$\pm$12.58&23.63$\pm$7.98&24.22$\pm$9.36&24.49$\pm$11.07&36.93$\pm$13.86&16.48$\pm$4.51&28.75$\pm$12.35&26.87$\pm$10.24&29.26$\pm$11.25\\
Yeast5&72.02$\pm$14.36&62.41$\pm$16.1&60.47$\pm$14.93&66.67$\pm$14.67&69.78$\pm$14.01&70.83$\pm$15.93&73.02$\pm$13.17&75.85$\pm$10.55&29.01$\pm$6.81&67.73$\pm$15.41&55.84$\pm$12.9&68.12$\pm$13.99\\
Aba.17vs789A&26.69$\pm$8.62&24.32$\pm$8.09&29.13$\pm$12.37&26.84$\pm$10.84&22.4$\pm$9.45&20.97$\pm$9.8&21.26$\pm$8.74&32.27$\pm$12.1&11.93$\pm$6.77&29.44$\pm$11.63&32.02$\pm$12.56&31.26$\pm$12.87\\
Aba.19vsABCD&6.55$\pm$8.76&7.3$\pm$6.77&10.82$\pm$9.99&5.21$\pm$5.63&7.11$\pm$5.82&8.08$\pm$7.93&5.99$\pm$5.94&5.79$\pm$4.97&5.56$\pm$4.89&7.73$\pm$5.45&7.62$\pm$3.47&8.28$\pm$6.28\\
Aba.21vs8&53.11$\pm$27.73&50.77$\pm$22.26&49.84$\pm$25.05&62.95$\pm$24.25&38.7$\pm$21.18&47.75$\pm$22.31&33.31$\pm$19.79&58.83$\pm$27.22&34.16$\pm$23.69&64.29$\pm$25.52&63.62$\pm$24.35&66.04$\pm$23.79\\
Aba.3vsB&100$\pm$0&100$\pm$0&100$\pm$0&100$\pm$0&100$\pm$0&100$\pm$0&100$\pm$0&99.75$\pm$1.77&95.48$\pm$6.71&100$\pm$0&99.99$\pm$0.1&100$\pm$0\\
Poker89vs5&5.11$\pm$3.71&3.76$\pm$3.92&5.47$\pm$4.38&5.73$\pm$3.9&10.67$\pm$8.74&12.96$\pm$10.04&14.81$\pm$10.17&14.68$\pm$11.68&2.59$\pm$4.74&14.62$\pm$13.22&15.33$\pm$12.45&14.4$\pm$12.67\\
Poker8vs6&67.42$\pm$24.75&60.38$\pm$26.96&59.71$\pm$25.71&63.65$\pm$26.26&67.31$\pm$28.43&64.95$\pm$24.32&81.6$\pm$20.7&37.06$\pm$32.01&3.05$\pm$6.71&98.57$\pm$5.79&92.05$\pm$12.45&96.53$\pm$13.66\\
Poker9vs7&39.58$\pm$33.22&12.86$\pm$23.86&53.58$\pm$29.71&77.96$\pm$26.25&68.72$\pm$23.11&63.32$\pm$27.29&57.52$\pm$27.27&25.44$\pm$24.37&18.17$\pm$21.83&58.15$\pm$33.9&62.05$\pm$30.98&63.67$\pm$31.56\\
Yeast6&47.17$\pm$18.78&52.04$\pm$13.19&49.05$\pm$16.96&58.25$\pm$16.49&34.71$\pm$13.85&36.55$\pm$14.75&34.39$\pm$13.14&60.22$\pm$17.37&16.86$\pm$4.35&48.08$\pm$17.37&53.76$\pm$13.2&48.42$\pm$14.9\\
WQ-W39vs5&10.35$\pm$9.43&13.25$\pm$10.47&13.34$\pm$10.78&13.55$\pm$9.76&9.93$\pm$9.01&6.49$\pm$5.4&8.98$\pm$8.21&16.89$\pm$12.56&10.86$\pm$10.55&12.48$\pm$14.84&10.76$\pm$12.8&12.46$\pm$13.28\\
Vowel0&99.98$\pm$0.08&99.97$\pm$0.2&99.99$\pm$0.04&99.98$\pm$0.14&99.98$\pm$0.12&100$\pm$0&99.99$\pm$0.06&95.06$\pm$3.84&79.47$\pm$8.85&100$\pm$0&93.62$\pm$0.1&100$\pm$0\\
Car&99.88$\pm$0.5&99.67$\pm$1.07&99.44$\pm$1.22&99.52$\pm$1.29&99.94$\pm$0.32&99.96$\pm$0.22&99.93$\pm$0.33&100$\pm$0&64.87$\pm$10.82&97.25$\pm$3.5&90.87$\pm$2.56&96.69$\pm$3.91\\
Shuttlec2vsc4&87.33$\pm$29.37&95.08$\pm$19.7&93.83$\pm$21.15&95.33$\pm$18.68&89.33$\pm$26.73&95.5$\pm$17.99&95.17$\pm$19.35&100$\pm$0&77.33$\pm$34.99&95.5$\pm$97.99&87$\pm$12.56&97$\pm$14.85\\
\hline
\textbf{Average}&52.42&51.51&51.56&56.43&51.85&53.85&51.98&55.63&36.1&58&57.10&\textbf{58.89}\\
\hline
\end{tabular}}
\end{table*}

\clearpage
\clearpage
\bibliographystyle{IEEEtranN}
\bibliography{refs}